\documentclass[11pt]{article}
\usepackage{geometry}                
\usepackage{setspace}
\usepackage[parfill]{parskip}    
\usepackage{newtxtext}

\usepackage{amsmath,amsfonts}
\usepackage{array}
\usepackage{textcomp}
\usepackage{stfloats}
\usepackage{verbatim}

\usepackage[utf8]{inputenc} 
\usepackage[T1]{fontenc}    
\usepackage{hyperref}       
\usepackage{url}            
\usepackage{booktabs}       
\usepackage{amsfonts}       
\usepackage{nicefrac}       
\usepackage{microtype}      
\usepackage{xcolor}         
\usepackage[ruled,vlined]{algorithm2e}
\usepackage{graphicx}
\usepackage{multirow}
\usepackage{subfigure}
\usepackage{wrapfig, blindtext}
\usepackage{amsmath}
\DeclareMathOperator*{\argmax}{arg\,max}

\usepackage{enumitem}
\usepackage{amsthm}
\newtheorem{theorem}{Theorem}

\newtheorem{lemma}{Lemma}
\newtheorem{definition}{Definition}

\usepackage{bm}
\usepackage[numbers,sort]{natbib}

\title{Spatial-Temporal-Fusion BNN: \\Variational Bayesian Feature Layer}
\author{Shiye Lei\thanks{S. Lei and Z. Tu are associated with School of Computer Science, Faculty of Engineering, the University of Sydney, Darlington NSW 2008, Australia and JD Explore Academy, JD.com Inc., Beijing 100176, China. This work was completed when they were interns at JD Explore Academy. Email: slei5230@uni.sydney.edu.au and zhtu3055@uni.sydney.edu.au.} \and 
Zhuozhuo Tu\footnotemark[1] \and
Leszek Rutkowski\thanks{L. Rutkowski is with the Department of Artificial Intelligence, University of Social Sciences, 90-113, Łódź, Poland, and also with the Systems Research Institute of the Polish Academy of Sciences, 01-447 Warsaw, Poland. Email:
lrutkowski@san.edu.pl.} \and
Feng Zhou\thanks{F. Zhou is associated with Department of Computer Science and Technology, BNRist Center, THU-Bosch Joint ML Center, Tsinghua University, Beijing 100086, China. Email: zhoufeng6288@tsinghua.edu.cn.} \and
Li Shen\thanks{L. Shen, F. He, and D. Tao are associated with JD Explore Academy, JD.com Inc., Beijing 100176, China. Email: shenli100@jd.com, hefengxiang@jd.com, and taodacheng@jd.com.} \and
Fengxiang He\footnotemark[4] \and
Dacheng Tao\footnotemark[4]
}
\date{}       

\begin{document}

\maketitle

\begin{abstract}
Bayesian neural networks (BNNs) have become a principal approach to alleviate overconfident predictions in deep learning, but they often suffer from scaling issues due to a large number of distribution parameters. In this paper, we discover that the first layer of a deep network possesses multiple disparate optima when solely retrained. This indicates a large posterior variance when the first layer is altered by a Bayesian layer, which motivates us to design a spatial-temporal-fusion BNN (STF-BNN) for efficiently scaling BNNs to large models: (1) first normally train a neural network from scratch to realize fast training; and (2) the first layer is converted to Bayesian and inferred by employing stochastic variational inference, while other layers are fixed. Compared to vanilla BNNs, our approach can greatly reduce the training time and the number of parameters, which contributes to scale BNNs efficiently. We further provide theoretical guarantees on the generalizability and the capability of mitigating overconfidence of STF-BNN. Comprehensive experiments demonstrate that STF-BNN (1) achieves the state-of-the-art performance on prediction and uncertainty quantification; (2) significantly improves adversarial robustness and privacy preservation; and (3) considerably reduces training time and memory costs.
\end{abstract}

\section{Introduction}
Neural networks often have significant uncertainty on their parameters in optimization \citep{rumelhart1985learning}. The point estimation of optimization to the parameters usually contributes to overconfident predictions \citep{hein2019relu}, which considerably undermines the trust of deep learning, especially in security-critical application domains, like medical diagnosis \citep{9353390} and autonomous driving \citep{huval2015empirical}. Bayesian neural networks (BNNs) are the main approaches to mitigate overconfidence by applying interval estimation. However, due to the tremendous number of trainable distribution parameters in BNNs, they experience issues in scaling to large-scale models and data, including prohibitively high convergence time, memory cost, application inflexibility, and training instability.

In this work, we explore the properties of solely retraining a single layer of the pretrained network, and the layer stability is defined by calculating the cosine similarity between parameters of the layer in the retrained net and in the pretrained net. Empirical studies on FashionMNIST and CIFAR-10 show that parameters of the last layer usually converge to similar results, while parameters of the first layer exhibit much lower stability, {\it i.e.}, more diverse convergence results after the retraining. The low layer stability in the first layer indicates that there exists multiple optimum with large discrepancy when the layer is retrained. Therefore, if the first layer with low stability is converted to Bayesian, the posterior distribution of the Bayesian layer tends to cover these dissimilar optimum, which induces a large posterior variance and helps alleviate overconfidence. 

\begin{figure*}[t]
    \centering
    \includegraphics[width=0.95\linewidth]{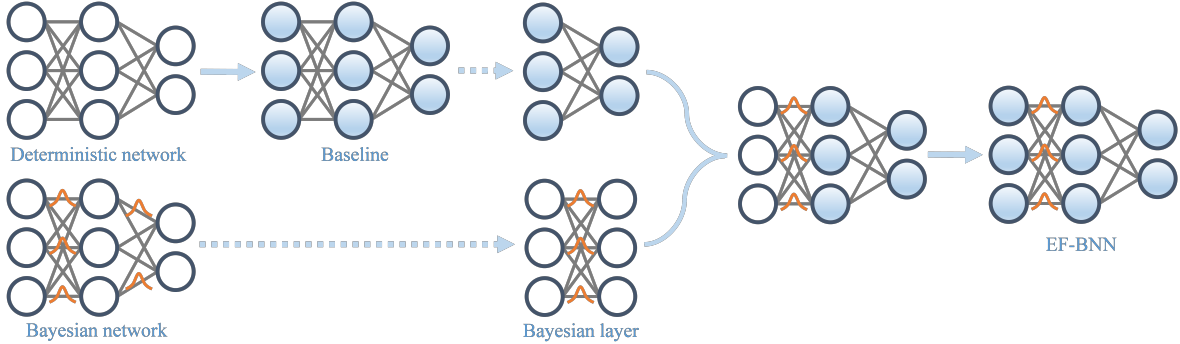}
    \caption{Spatial-temporal-fusion BNN architecture. The lines between circles denote the network parameters, and Bayesian layers are marked by orange bell curves. Hollow circles are colored by training, which is denoted by the arrows with blue solid lines.
    }
    \label{figure: stf bnn architecture}
\end{figure*}

Inspired by the findings in the experiments {\it w.r.t.} layer stability, we propose the {\it spatial-temporal-fusion BNN} (STF-BNN) that realizes the fusion of optimization and Bayesian inference for training neural networks from both “spatial” and “temporal” aspects, and the schematic diagram of the STF-BNN is presented in Figure \ref{figure: stf bnn architecture}. In the first training phase, we employ optimization methods to train the whole neural network to reach fast and memory-efficient learning. During the second training phase, we employ Bayesian inference to train the first layer, while all other layers are frozen. Due to the tremendous decrease in the number of distribution parameters compared to vanilla BNNs, Bayesian inference achieves fast convergence in this phase. Then, we theoretically prove the generalization ability and the capability of mitigating overconfidence of our STF-BNN.

Comprehensive experiments of two standard neural network architectures, VGG-19 and Wide ResNet-28-10, are conducted on four benchmark datasets, CIFAR-10, CIFAR-100, CIFAR-10-C, and CIFAR-100-C, to evaluate STF-BNNs from three aspects: uncertainty quantification, privacy protection, and adversarial robustness. Empirical results show that STF-BNNs can efficiently improve performance in these aspects. Besides, the performance of STF-BNNs in uncertainty quantification is on-par with other state-of-the-art approaches, while significantly reducing training time and memory costs.

In sum, the main contributions are as follows:
\begin{itemize}
    \item For the first time in literature we reveal the large disparity of layer stability between different layers in deep neural networks, which prompts us to design the STF-BNN;
    \item Based on spatial and temporal fusions, we propose the STF-BNN, which can be trained fast and also reduces the memory costs;
    \item We theoretically prove the generalization performance and the capability of mitigating overconfidence of STF-BNN;
    \item Comprehensive experiments are conducted on STF-BNNs and show its superior performance on uncertainty quantification, privacy preservation, and adversarial robustness.
\end{itemize}

\section{Related Works}
\textbf{Bayesian neural networks.} From the Bayesian perspective, the parameters of neural networks obey a specific distribution, which terms the posterior distribution and is approximated by Bayesian inference. A famous Bayesian inference approach is Markov chain Monte Carlo (MCMC) \citep{hastings1970monte, gamerman2006markov, neal2011mcmc}, which constructs a Markov chain that performs a Monte Carlo sampler to approximate the posterior. To leverage MCMC on training deep Bayesian neural networks, stochastic gradient MCMC (SGMCMC) employs stochastic gradients to speed up and scale MCMC \citep{7961239}, such as Stochastic Gradient Langevin Dynamic (SGLD) \citep{welling2011bayesian}, Stochastic Gradient Hamiltonian Monte Carlo (SGHMC) \citep{chen2014stochastic}, Stochastic Gradient Fisher Scoring (SGFS) \citep{ahn2012bayesian}, Cyclical SGMCMC \citep{zhang2020cyclical}, and a summary \citep{ma2015complete}. Another major Bayesian inference method is variational inference (VI) \citep{blei2017variational}, which casts inference as an optimization problem and aims to maximize the evidence lower bound (ELBO). Stochastic variational inference (SVI) \citep{graves2011practical, hoffman2013stochastic, 6937190} also adopts stochastic gradients to speed up and scale VI. Bayes by backprop \citep{blundell2015weight} and Radial BNNs \citep{farquhar2020radial} provide a practical solution in implementations. \citet{khan2018fast} further speed up SVI based on Adam \citep{kingma2014adam}. \citet{krishnan2020specifying} and \citet{deng2020bayesadapter} also utilize the weights of pretrained neural networks as the initial posterior mean in training BNNs to improve the convergence performance.

\textbf{Uncertainty quantification.} Considering the overconfident predictions given by deep neural networks \citep{guo2017calibration}, quantifying the uncertainty of model predictions is crucial for improving the explainability and reliability of networks. Some metrics have been proposed for uncertainty quantification \citep{gustafsson2020evaluating,huang2019evaluating}, such as expected calibration error (ECE) that measures the difference between the prediction accuracy and the predictive confidence score produced by networks \citep{guo2017calibration}. Many approaches are also proposed to better estimate the prediction uncertainty by designing novel architectures, loss functions, and training strategies based on deterministic networks. SDE-Net \citep{kong2020sde} is consisted of a drift net and a diffusion net to fit the training examples and produce the uncertainty, respectively. \citet{van2020uncertainty} propose a novel loss function that inspects the centroid of data to improve the capability of uncertainty quantification. \citet{liu2020simple} also achieve this by adding a weight normalization step during training and replacing the output layer with a Gaussian Process. Except for the approaches that employ deterministic neural networks for estimating the prediction uncertainty via single forward propagation, there are many ensemble methods, such as deep ensemble \citep{lakshminarayanan2017simple} and Batch ensemble \citep{wen2020batchensemble}, which also designed for uncertainty quantification.
Because their parameters are equipped with distributions, BNNs have gradually become dominant approaches in quantifying the uncertainty of neural networks \citep{gal2016dropout,dusenberry2020efficient} in recent years. \citet{kristiadi2020being} prove that equipping the last linear classifier with a Bayesian layer can mitigate the overconfidence of ReLU networks, while our approach introduces the Bayesian layer into the first non-linear layer with two-phase training for more efficient training.

\textbf{Privacy protection and adversarial robustness.} Privacy and security concerns in deep learning are rising as deep learning has been applied to increasingly sensitive domains \citep{dwork2008differential,dwork2014algorithmic,madry2018towards}. Since neural networks easily ``memorize'' the training data, attackers can infer the sensitive information contained in the training data by inspecting the well-trained networks \citep{shokri2017membership}. For example, a membership inference (MI) attack aims to infer whether an example is in the training data or not. Some works have also proposed to mitigate the privacy leakage in deep learning models through injecting noise into gradients during the training process \citep{mironov2017renyi,he2020tighter,abadi2016deep,9115837}. 

In addition to the risk of privacy leakage, neural networks have also been reported to be vulnerable to adversarial attacks \citep{madry2018towards,8611298}: networks sometimes output completely different predictions when the input images are slightly modified. The adversarial attack is firstly probed by \citet{43405}, which propose the fast gradient sign method (FGSM) to generate adversarial examples by single gradient ascent. \citet{madry2018towards} leverage the projection gradient descent (PGD) on the adversarial attack. Compared to FGSM, PGD generates higher-quality adversarial examples via multiple iterations of gradient ascent. \citet{carlini2017towards} develop three adversarial attacks, which make the perturbations quasi-imperceptible by restricting their $l_0$, $l_2$, and  $l_\infty$ norms. A common practice for improving adversarial robustness is adversarial training \citep{shafahi2019adversarial} which applies a minimax approach to robust neural networks. Some methods also modify optimization \citep{ye2018bayesian} and network architectures \citep{liu2018adv} to improve the robustness. However, \citet{zhang2019theoretically} show the trade-off between adversarial robustness and prediction accuracy and claim that adversarial training leads to underfitting on the normal sample and hurts model accuracy. Furthermore, they mitigate the trade-off between robustness and accuracy by disentangling the robust error into natural error and boundary error. \citet{zhang2021geometryaware} also alleviate the trade-off by assigning small weights to non-important samples for saving model capacity and developing a reweighting-based framework that improves robustness while preserving accuracy.

\section{Preliminaries}
We denote the training set by $\mathcal{S}=\{(\mathbf{x}_i, y_i)\}_{i=1}^m$, where $\mathbf{x}_i\in \mathbb{R}^n$, $n$ is the dimension of input data, $y_i\in \{1,\dots,k\}$, $k$ is the number of classes, and $m=|\mathcal{S}|$ is the training sample size. We assume that $(\mathbf{x}_i, y_i)$ are independent and identically distributed (i.i.d.) random variables drawn from the data generating distribution $\mathcal{D}$. Denote the ReLU neural network as $f_{\boldsymbol{\theta}}(\mathbf{x})=W_{d} \phi\left(W_{d-1} \phi\left(\cdots \phi\left(W_{1} \mathbf{x}\right)\right)\right): \mathbb{R}^n \rightarrow \mathbb{R}^k$, which is a $d$ layer feed-forward network parameterized by $\boldsymbol{\theta}=\text{vec}(\{W_i\}^d_{i=1})$, here $\phi$ denotes the ReLU activation function.
The logit $f_{\boldsymbol{\theta}}(\mathbf{x})$ is normalized by the non-linear Softmax function $o(f_{\boldsymbol{\theta}}(\mathbf{x}))=\operatorname{softmax}(f_{\boldsymbol{\theta}}(\mathbf{x}))$, and $o(f_{\boldsymbol{\theta}}(\mathbf{x}))$ is assumed to be a discrete probability density function. Let $o^{(i)}(f_{\boldsymbol{\theta}}(\mathbf{x}))$ be the $i$-th component of $o(f_{\boldsymbol{\theta}}(\mathbf{x}))$, hence $\sum^k_{i=1}o^{(i)}(f_{\boldsymbol{\theta}}(\mathbf{x})) = 1$. We define $\hat{p}(\mathbf{x})= o^{(y)}(f_{\boldsymbol{\theta}}(\mathbf{x}))$ and $\hat{y}(\mathbf{x})=\argmax_{i}o^{(i)}(f_{\boldsymbol{\theta}}(\mathbf{x}))$ to denote the confidence score and the predicted labels by the network $f_{\boldsymbol{\theta}}$ on $\mathbf{x}$, respectively.

For the classical setting that $f_{\boldsymbol{\theta}}$ is a deterministic neural network, a maximum likelihood estimation ${\boldsymbol{\theta}}_{\operatorname{MLE}}=\argmax_{\boldsymbol{\theta}}\log \Pr\left[\hat{y}(\mathbf{x}_i)=y_i| (\mathbf{x}_i,y_i)\in \mathcal{S}\right]$ is returned by optimizing $f_{\boldsymbol{\theta}}$ on $\mathcal{S}$ in practice. However, due to the randomness of the learning algorithm, let $\mathbb{Q}(\boldsymbol{\theta})$ denote the posterior distribution returned by the training algorithm leveraged on $\mathcal{S}$, and ${\boldsymbol{\theta}}_{\operatorname{MLE}}\sim\mathbb{Q}(\boldsymbol{\theta})$.

If $f_{\boldsymbol{\theta}}$ is a Bayesian neural network, the posterior $\mathbb{Q}(\boldsymbol{\theta})$ is often approximated by variational inference, which aims to maximize the evidence lower bound (ELBO):
\begin{equation}
    \operatorname{ELBO}=\mathbb{E}_{\boldsymbol{\theta}\sim\mathbb{Q}(\boldsymbol{\theta})}\left(\log\mathbb{P}\left(\mathcal{S}|\boldsymbol{\theta}\right)\right) - \operatorname{KL}\left(\mathbb{Q}\|\mathbb{P}\right),
\end{equation}
where the likelihood $\log\mathbb{P}\left(\mathcal{S}|\boldsymbol{\theta}\right)\approx\sum_{i=1}^m\log o^{(y_i)}(\mathbf{x}_i)$ and $\operatorname{KL}\left(\mathbb{Q}\|\mathbb{P}\right)$ is the KL-divergence between the approximate posterior $\mathbb{Q}(\boldsymbol{\theta})$ and the prior $\mathbb{P}(\boldsymbol{\theta})$:
\begin{equation}
{\rm KL}(\mathbb{Q} \| \mathbb{P})=\mathbb{E}_{\boldsymbol{\theta} \sim \mathbb{Q}}\left(\log \frac{\mathbb{Q}(\boldsymbol{\theta})}{\mathbb{P}(\boldsymbol{\theta})}\right).
\end{equation}
In this paper, we adopt Gaussian distribution as the approximate posterior, {\it i.e.}, $\mathbb{Q}(\boldsymbol{\theta})=\mathcal{N}(\boldsymbol{\theta}|\boldsymbol{\mu}, \boldsymbol{\Sigma})$, where $\boldsymbol{\mu}$ is the mean and $\boldsymbol{\Sigma}$ is the covariance matrix.

To estimate the probability of the output for a BNN by the posterior mean and variance of its parameters, we introduce Lemma \ref{lemma:1} as follows.
\begin{lemma}[\citet{mackay1992evidence}]
\label{lemma:1}
For a binary classification BNN $f_{\boldsymbol{\theta}}: \mathbb{R}^{n} \rightarrow \mathbb{R}$, where $\boldsymbol{\theta}$ is the parameters of $f_{\boldsymbol{\theta}}$ and is sampled from the Gaussian distribution $\mathcal{N}(\boldsymbol{\theta} | \boldsymbol{\mu}, \boldsymbol{\Sigma})$, and $\mathbf{d}(\mathbf{x}):=\left.\nabla_{\boldsymbol{\theta}} f_{{\boldsymbol{\theta}}}(\mathbf{x})\right|_{{\boldsymbol{\theta}=\boldsymbol{\mu}}}$. The sigmoid function is defined as $\tau(z)=1/(1+\exp(-z))$ for $z\in \mathbb{R}$. Then we have that
\begin{equation}
\label{eq:approximately equal}
    \operatorname{Pr}[y=1|\mathbf{x}] \approx \tau(z(\mathbf{x})),
\end{equation}
where
\begin{equation}
z({\mathbf{x}}):=\frac{f_{\boldsymbol{\mu}}(\mathbf{x})}{\sqrt{1+\pi / 8 \mathbf{d}^{\top} {\boldsymbol{\Sigma}} \mathbf{d}}}.
\label{eq:zx}
\end{equation}
\end{lemma}
The approximately equal in Eq. \ref{eq:approximately equal} comes from the first-order Taylor expansion of $f_{\boldsymbol{\theta}}$ at $\boldsymbol{\mu}$ \citep{mackay1995probable}: $f_{\boldsymbol{\theta}}(\mathbf{x}) \approx f_{\boldsymbol{\mu}}(\mathbf{x})+\mathbf{d}(\mathbf{x})^{\top}(\boldsymbol{\theta}-\boldsymbol{\mu})$. In Lemma \ref{lemma:1}, $z(\mathbf{x})$ can be regarded as the logit, and $\tau(z(\mathbf{x}))$ is the probability or the confidence score for the binary BNN.

\section{Layer Stability}
\label{sec:layer uncertainty}

\begin{wrapfigure}{r}{0.43\textwidth}
\vspace{-4mm}
\begin{minipage}{0.43\textwidth}
            \centering
    		\includegraphics[width=0.48\columnwidth]{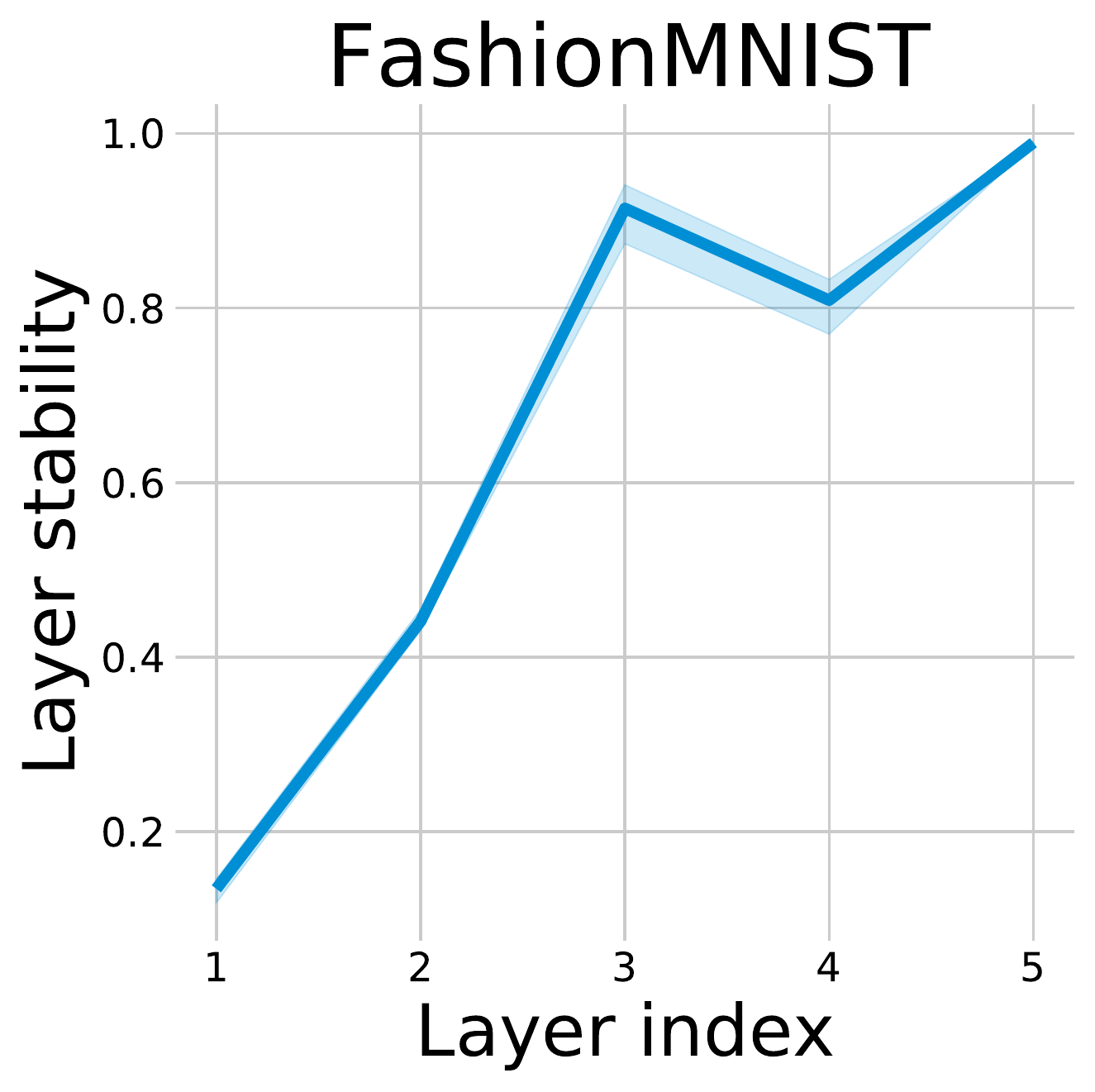}
   		 	\includegraphics[width=0.48\columnwidth]{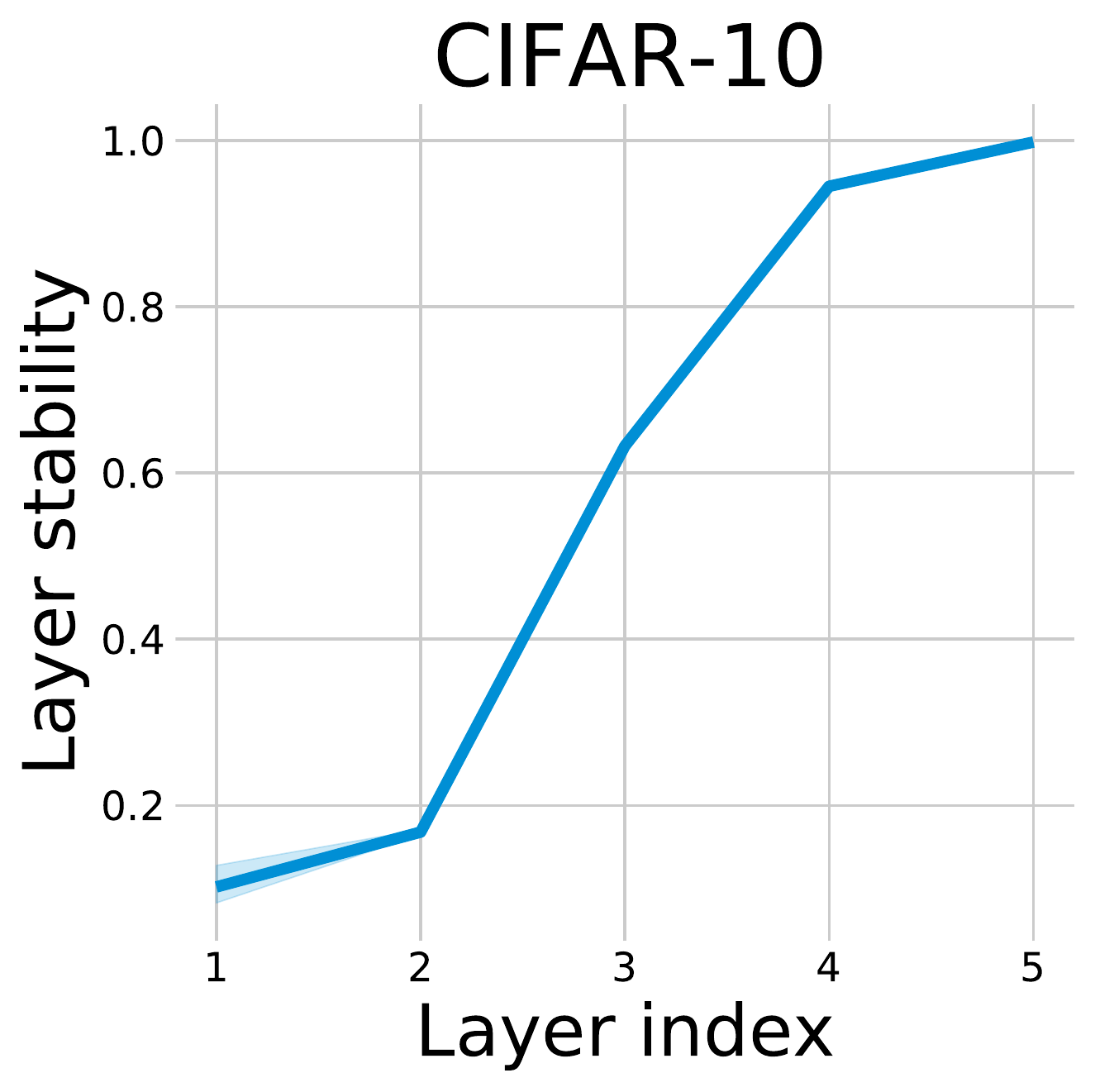}
    		\end{minipage}
\vspace{-2mm}
\caption{Layer stability as a function of layer index on FashionMNIST and CIFAR-10; the darker lines show the average over five seeds and the shaded area shows the standard deviations.}
\label{figure:layer variability}
\end{wrapfigure}

In this section, we explore the parameter stability of different layers in neural networks, which sheds light on designing following spatial-temporal-fusion BNN. In detail, a neural network $f_{pre}$ is first trained on the training set $\mathcal{S}$. Then, the $k$-th layer of $f_{pre}$ is initialized and retrained on $\mathcal{S}$ while the parameters of other layers are fixed until the training process has converged, so we get the retrained network $f_{re\text{-}k}$. With $f_{re\text{-}k}$, the stability of the $k$-th layer of $f_{pre}$ can be defined as below.
\begin{definition}[Layer stability]
\label{definition:layer stability}
Let $W, W' \in \mathbb{R}^{a\times b}$ denote the weight matrix of the $k$-th layer of the pretrained network $f_{pre}$ and the retrained network $f_{re\text{-}k}$, respectively. $a$ and $b$ are the dimensions of the output and the input of the $k$-the layer. Then, the stability of the $k$-th layer of $f_{pre}$ is defined as
\begin{equation}
\label{eq:stability}
    \frac{1}{a}\sum_{i=1}^a \frac{\vert \mathbf{w}^{\top}_i \mathbf{w}'_{i} \vert}{\Vert \mathbf{w}_i \Vert \Vert \mathbf{w}'_{i} \Vert},
\end{equation}
where $\mathbf{w}_i,\mathbf{w}'_i \in \mathbb{R}^{b}$ denote the $i$-th row vector of $W_i$ and $W'_i$, respectively, for $i\in[1,a]$.
\end{definition}

We discard the Euclidean distance $\vert\mathbf{w}_i-\mathbf{w}^\prime_i\vert$ 
in the definition because ReLU networks possess the scale-invariant property, which states that the output of the ReLU network is invariant when the parameters of the layer are multiplied by a positive constant $c$ and the logits are divided by $c$. Therefore, scaling parameters of a single layer has no effect on final predictions, and thus a series of scaled weights should be treated equally. The cosine similarity is employed to measure the layer stability because it is non-sensitive to this scaling. According to Definition \ref{definition:layer stability}, smaller layer stability indicates $\mathbf{w}$ and $\mathbf{w}'$ become more orthogonal, and thus 
they employ a more dissimilar ratio of the previous output to shape new features. 
It is worth noting that the definition of layer stability is different from the co-adaptation \citep{hinton2012improving,cogswell2015reducing}, which measures the correlations between activations, while the layer stability focuses on the network parameters.

To measure the layer stability, we first trained $5$ five-hidden-layer MLPs, {\it i.e.}, $f_{pre}$ on FashionMNIST \citep{xiao2017/online} and CIFAR-10 \citep{krizhevsky2009learning}, respectively, and $f_{re\text{-}k}$ are subsequently obtained by retraining the specific layer of $f_{pre}$ on the same training set. Then, according to Eq. \ref{eq:stability}, we can compute the $k$-th layer's stability by comparing the parameters of the $k$-th layer in $f_{pre}$ and in $f_{re\text{-}k}$, and the stability of different layers is plotted in Figure \ref{figure:layer variability}. From the figure, we have two observations: (1) the first layer has the lowest stability compared to the subsequent layers; and (2) the stability of the last layer is close to one. 

The experiment reveals the large disparity of layer stability among different layers in deep neural networks. Moreover, low layer stability shows that there exists multiple optima with large discrepancies when the layer is retrained. Therefore, if we convert the layers with low layer stability into Bayesian layers, the posterior distribution of Bayesian layers tends to cover these dissimilar optima, which induces a large posterior variance and hence helps mitigate overconfidence.

\section{Methods}
\label{sec:methods}
Inspired by the finding that solely retraining the first layer of the pretrained networks induces low layer stability, we construct the spatial-temporal-fusion BNN by making the first layer as Bayesian to efficiently mitigate overconfidence with faster training and less memory cost.

\subsection{Spatial-Temporal Fusion}
Spatial-temporal-fusion BNNs (STF-BNNs) realize the fusion 
from both spatial and temporal aspects. From the spatial perspective, STF-BNNs fuse the conventional deterministic neural networks with the Bayesian layer, which allows the network to capture the parameter uncertainty and also substantially reduce the number of distribution parameters compared to vanilla BNNs. From the temporal perspective, the fusion of optimization and Bayesian inference makes the training of STF-BNNs more efficient. The training procedure of STF-BNNs can be divided into two phases as follows.

During the first training phase, STF-BNNs employ optimization methods, such as stochastic gradient descent (SGD), to train the whole network $f_{\boldsymbol{\theta}}$ to $f_{{\boldsymbol{\theta}}_{\operatorname{MLE}}}$. In this way, one can achieve fast training of the network.

During the second training phase, STF-BNNs adopt Bayesian inference to train the first layer of the network $f_{{\boldsymbol{\theta}}_{\operatorname{MLE}}}$, in order to quantify uncertainty for improving the reliability in a relatively cost. 
The parameter ${\boldsymbol{\theta}}_{\operatorname{MLE}}$ is partitioned into ${\boldsymbol{\theta}}_1$ and ${\boldsymbol{\theta}}_2$: ${\boldsymbol{\theta}}_1$ denotes the parameters of the first layer, and ${\boldsymbol{\theta}}_2$ is the parameters of the subsequent layers. Then we parameterize the first layer with the posterior mean and variance of weights by ${\boldsymbol{\theta}}_1 \sim\mathcal{N} (\boldsymbol{\mu}_1,\boldsymbol{\Sigma}_1)$, and Bayesian inference is employed to infer the approximate posterior $\mathcal{N}(\boldsymbol{\mu}_1,\boldsymbol{\Sigma}_1)$ while ${\boldsymbol{\theta}}_2$ is fixed. In this training phase, {\it Bayes by backprop} \citep{blundell2015weight} is employed for practical implementation. The entire process of training an STF-BNN is presented in Algorithm \ref{algorithm:1}, and we employ the trainable parameter $\boldsymbol{\rho}$ to generate the covariance matrix $\boldsymbol{\Sigma}_1$ for guaranteeing the positive variance.
\begin{algorithm}[t]
\label{algorithm:1}
\SetAlgoLined
Optimize a deterministic neural network $f_{{\boldsymbol{\theta}}_{\operatorname{MLE}}}$\;

Partition ${{\boldsymbol{\theta}}_{\operatorname{MLE}}}$ into ${\boldsymbol{\theta}}_1$ and ${\boldsymbol{\theta}}_2$, and reinitialize ${\boldsymbol{\theta}}_1$ with variational parameters $\boldsymbol{\gamma}=(\boldsymbol{\mu}_1, \boldsymbol{\rho})$\;

 \For{epoch in num\_epochs}{
Sample $\boldsymbol{\epsilon}\sim \mathcal{N}(0,I)$ and $\boldsymbol{\theta}_1$:
  \begin{equation}
  \begin{aligned}
    \boldsymbol{\Sigma}_1 &= \log (1+\exp (\boldsymbol{\rho})) \\
    \boldsymbol{\theta}_1 &= \boldsymbol{\mu}_1 + \boldsymbol{\Sigma}_1 \circ \boldsymbol{\epsilon};\;
  \end{aligned}
  \end{equation}

Compute $-\operatorname{ELBO}$ $l(\boldsymbol{\theta}_1, \boldsymbol{\gamma})$ with batch size $M$:
  \begin{equation}
      l(\boldsymbol{\theta}_1, \boldsymbol{\gamma})=\sum_{i=1}^M \mathcal{L}( f(\mathbf{x}_i | \boldsymbol{\theta}_1, \boldsymbol{\gamma}), y_i) + \operatorname{KL}(\mathbb{Q}\Vert \mathbb{P}),\; 
  \end{equation}
  where $\mathcal{L}$ is the cross-entropy loss, and $\operatorname{KL}(\mathbb{Q}\Vert \mathbb{P})$ is the KL-divergence between the approximate posterior $\mathbb{Q}(\boldsymbol{\theta}_1)=\mathcal{N}(\boldsymbol{\mu}_1,\boldsymbol{\Sigma}_1)$ and the prior $\mathbb{P}(\boldsymbol{\theta}_1)=\mathcal{N}(0, I)$.

~\\

Calculate gradients of variational parameters:
  \begin{equation}
\begin{aligned}
\nabla_{\boldsymbol{\mu}_1}l &=\frac{\partial l(\boldsymbol{\theta}_1, \boldsymbol{\gamma})}{\partial \boldsymbol{\theta}_1}+\frac{\partial l(\boldsymbol{\theta}_1, \boldsymbol{\gamma})}{\partial \boldsymbol{\mu}_1}, \\
\nabla_{\boldsymbol{\rho}}l &=\frac{\partial l(\boldsymbol{\theta}_1, \boldsymbol{\gamma})}{\partial \boldsymbol{\theta}_1} \frac{\boldsymbol{\epsilon}}{1+\exp (-\boldsymbol{\rho})}+\frac{\partial l(\boldsymbol{\theta}_1, \boldsymbol{\gamma})}{\partial \boldsymbol{\rho}};\;
\end{aligned}
\end{equation}

Update the variational parameters: 
\begin{equation}
\begin{array}{l}
\boldsymbol{\mu}_1 \leftarrow \boldsymbol{\mu}_1-\alpha \nabla_{\boldsymbol{\mu}_1}l \\
\boldsymbol{\rho} \leftarrow \boldsymbol{\rho}-\alpha \nabla_{\boldsymbol{\rho}}l;\;
\end{array}
\end{equation}
 }
\caption{Training procedure of the STF-BNN}
\end{algorithm}

\subsection{Generalization Bound of STF-BNN}
In this section, we explore the generalizability \citep{mohri2012foundations,he2020recent} for the STF-BNN based on the PAC-Bayes generalization bound \citep{mcallester1999some}.

The parameters $\boldsymbol{\theta}$ of the STF-BNN are partitioned into ${\boldsymbol{\theta}}_1$ and ${\boldsymbol{\theta}}_2$ according to whether they are updated in Bayesian inference or optimization, respectively. For ${\boldsymbol{\theta}}_1\sim\mathcal{N}(\boldsymbol{\mu}_1,\boldsymbol{\Sigma}_1)$, we assume the approximate posterior $\mathbb{Q}(\boldsymbol{\theta}_1)=\mathcal{N}(\boldsymbol{\mu}_1,\boldsymbol{\Sigma}_1)$ obeys mean-field assumption, which implies ${\boldsymbol{\Sigma}}_{1}$ is a diagonal matrix. On the other hand, $\boldsymbol{\theta}_2$ is optimized by SGD. Without loss of generality, we assume $\mathbb{E}_{\boldsymbol{\theta}_2\sim\mathbb{Q}(\boldsymbol{\theta}_2)}[\boldsymbol{\theta}_2]=\boldsymbol{0}$. Then, $\boldsymbol{\theta}_2$ has an analytic stationary distribution according to \citet{mandt2017stochastic}:
\begin{equation}
\label{eq:distribution theta2}
\mathbb{Q}(\boldsymbol{\theta}_2)=\frac{1}{\sqrt{2 \pi \operatorname{det}(\boldsymbol{\Sigma}_{2})}} \exp \left\{-\frac{1}{2} \boldsymbol{\theta}_2^{\top} \boldsymbol{\Sigma}_{2}^{-1} \boldsymbol{\theta}_2\right\}.
\end{equation}

$\boldsymbol{\Sigma}_2=\operatorname{cov}(\boldsymbol{\theta}_2, \boldsymbol{\theta}_2)$ is the covariance matrix of $\boldsymbol{\theta}_2$, and $\boldsymbol{\Sigma}_{12}={\rm cov}(\boldsymbol{\theta}_1, \boldsymbol{\theta}_2)$ denotes the covariance matrix between $\boldsymbol{\theta}_1$ and $\boldsymbol{\theta}_2$.

$0-1$ loss is employed in the theoretical analysis, and the expected risk and the empirical risk of STF-BNN are defined as:
\begin{equation}
    \mathcal{R}(\mathbb{Q}) = \mathbb{E}_{(\mathbf{x},y)\sim \mathcal{D}}\mathbb{E}_{(\boldsymbol{\theta}_1,\boldsymbol{\theta}_2)\sim \mathbb{Q}(\boldsymbol{\theta}_1,\boldsymbol{\theta}_2)} \left[\mathbf{1}\left(y\neq \hat{y}\left(\mathbf{x}\right)\right)\right]
\end{equation}
\begin{equation}
    \hat{\mathcal{R}}(\mathbb{Q}) 
    =  \frac{1}{m}\sum_{i=1}^m\mathbb{E}_{(\boldsymbol{\theta}_1,\boldsymbol{\theta}_2)\sim \mathbb{Q}(\boldsymbol{\theta}_1,\boldsymbol{\theta}_2)} \left[ \mathbf{1}\left(y_i\neq \hat{y}\left(\mathbf{x}_i\right)\right)\right]
\end{equation}
respectively, where $\mathbf{1}(\cdot)$ is the indicator function, $\mathbb{Q}(\boldsymbol{\theta}_1,\boldsymbol{\theta}_2)$ is the joint posterior distribution for $\boldsymbol{\theta}_1$ and $\boldsymbol{\theta}_2$. Then, we obtain a generalization bound for STF-BNNs as follows:

\begin{theorem}[Generalization bound of STF-BNN]
\label{thm:generalization bound}
For any positive real $\delta\in (0,1)$, with probability at least $1-\delta$ over a training sample set of size $m$, we have the following inequality for the distribution of the output hypothesis function $\mathbb{Q}(\boldsymbol{\theta}_1,\boldsymbol{\theta}_2)$:
\begin{equation}
\begin{aligned}
\mathcal{R}(\mathbb{Q}) \leq  \hat{\mathcal{R}}(\mathbb{Q})+
\sqrt{\frac{\Delta  +2\log \frac{1}{\delta}+2\log m+4}{4 m-2}},
\end{aligned}
\end{equation}
where
\begin{equation}
\begin{aligned}
    \Delta = & 2{\rm KL}(\mathbb{Q}(\boldsymbol{\theta}_1) \| \mathbb{P}(\boldsymbol{\theta}_1)) + \operatorname{tr}(\boldsymbol{\Sigma}-2I) \\ & - \log(\det(\boldsymbol{\Sigma})) + \|\boldsymbol{\Sigma}_{1}\|^2 + \|\boldsymbol{\Sigma}_{2}\|^2,
\end{aligned}
\end{equation}
and $\boldsymbol{\Sigma}$ is the covariance matrix of $\boldsymbol{\theta}$ and is defined as
\begin{equation}
    \boldsymbol{\Sigma}={\rm cov}(\boldsymbol{\theta}, \boldsymbol{\theta})=\left[\begin{array}{ll}
\boldsymbol{\Sigma}_{1} & \boldsymbol{\Sigma}_{12} \\
\boldsymbol{\Sigma}_{12}^{\top} & \boldsymbol{\Sigma}_{2}
\end{array}\right].
\end{equation}
\end{theorem}
The proof for this generalization bound has two parts: (1) disentangle the KL divergence {\it w.r.t.} $\boldsymbol{\theta}$ to the KL divergence {\it w.r.t.} $\boldsymbol{\theta}_1$ and $\boldsymbol{\theta}_2$, respectively; and (2) adopt the PAC-Bayesian framework to obtain the bound. A detailed proof is given in Section \ref{sec:proof1}.

\subsection{STF-BNN Mitigate Overconfidence}
For the binary classification STF-BNN $f_{\boldsymbol{\theta}_1, \boldsymbol{\theta}_2}: \mathbb{R}^n\rightarrow\mathbb{R}$, we theoretically study its capability in tackling the overconfidence in this section. 
According to Lemma \ref{lemma:1},  
by defining $\mathbf{d}_1(\mathbf{x}):=\left.\nabla_{\boldsymbol{\theta}_1} f_{{\boldsymbol{\theta}_1, \boldsymbol{\theta}_2}}(\mathbf{x})\right|_{\boldsymbol{\theta}_1={\boldsymbol{\mu}_1}}$ for the binary STF-BNN $f_{\boldsymbol{\theta}_1, \boldsymbol{\theta}_2}$, where $\boldsymbol{\theta}_1$ has the approximated posterior $\mathcal{N}(\boldsymbol{\theta}_1|\boldsymbol{\mu}_1, \boldsymbol{\Sigma}_{1})$, we can estimate the logit of $f_{\boldsymbol{\theta}_1, \boldsymbol{\theta}_2}$ as follows:
\begin{equation}
\label{eq:ef bnn zx}
z(\mathbf{x}):= \frac{f_{{\boldsymbol{\mu}_1}, \boldsymbol{\theta}_2}(\mathbf{x})}{\sqrt{1+\pi / 8 \mathbf{d}_{1}^{\top} {\boldsymbol{\Sigma}_{1}} \mathbf{d}_{1}}}.
\end{equation}

\begin{theorem}[STF-BNN mitigates overconfidence]
Let $f_{\boldsymbol{\theta}_1, \boldsymbol{\theta}_2}: \mathbb{R}^n \rightarrow \mathbb{R}$ be a binary ReLU STF-BNN parameterized by $\boldsymbol{\theta}_1\in \mathbb{R}^{r\times n}$ and $\boldsymbol{\theta}_2$, where $r$ is the output dimension of the first Bayesian layer in the STF-BNN. Let $\mathcal{N}(\boldsymbol{\theta}_1|\boldsymbol{\mu}_1,\boldsymbol{\Sigma}_1)$ be the approximated posterior over $\boldsymbol{\theta}_1$, and $\boldsymbol{\theta}_2$ is obtained by the optimization. Then for any input $\mathbf{x}\in\mathbb{R}^{n}$,
\begin{equation}
\lim _{\delta \rightarrow \infty} \tau(|z(\delta \mathbf{x})|) \leq \tau\left(\frac{\|\mathbf{u}\| \|\mathbf{w}\|}{s_{\min}(\mathbf{u}^{\top}\mathbf{J}) \sqrt{\pi / 8 \lambda_{\min }(\boldsymbol{\Sigma}_1)}}\right),
\end{equation}
where $\mathbf{w}\in \mathbb{R}^{r\times n}$ and $\mathbf{u}\in \mathbb{R}^r$ depend only on $\boldsymbol{\mu}_1$ and $\boldsymbol{\theta}_2$, respectively, the matrix $\mathbf{J}:=\left.\frac{\partial \mathbf{w}}{\partial \boldsymbol{\theta}_1}\right|_{\boldsymbol{\mu}_1}$ is the Jacobian of $\mathbf{w}$ w.r.t. $\boldsymbol{\theta}_1$ at $\boldsymbol{\mu}_1$, and functions $\lambda_{\min} (\cdot)$ and ${s}_{\min} (\cdot)$ return the minimum eigenvalue and singular value of their matrix argument, respectively.
\label{thm:overconfidence}
\end{theorem}
The proof is given in Section \ref{sec:proof2}. Because $\tau(z(\mathbf{x}))$ is the confidence score of $\mathbf{x}$, Theorem \ref{thm:overconfidence} states that the confidence score of $\delta \mathbf{x}$, which is an input far away from the training data when $\delta\rightarrow\infty$, can be bounded and will not be close to zero or one in our STF-BNN. Therefore, our approach theoretically limits the confidence score of inputs far away from the training data and mitigates the problem of overconfidence.
\section{Experiments}
\label{sec:experiments}
In this section, we analyze the properties of STF-BNN through a comprehensive empirical study on three tasks: uncertainty quantification, privacy preservation, and adversarial robustness.

\subsection{Setup}
In our experiments, We utilize two popular network architectures of VGG-19 \citep{simonyan2014very} and Wide ResNet-28-10 \citep{zagoruyko2016wide} on four datasets: CIFAR-10, CIFAR-100 \citep{krizhevsky2009learning}, and their corresponding corruptions CIFAR-10-C and CIFAR-100-C \citep{hendrycks2019robustness}. The preprocessing of a dataset and data augmentation of training images follow \citet{zagoruyko2016wide}.

We optimize deterministic models for $200$ epochs in the first training phase. The learning rate is initialed as $0.1$ and is decayed by $0.2$ at the $[60, 120, 160]$ epoch. Weight decay factor is set as $5e-4$. The networks only undergo the first training phase are termed as {\it baselines}. In the second training phase, we train our STF-BNNs for $30$ epochs by variational inference. Learning rate is initialed as $0.1$ and is decayed by $0.2$ at the $[10, 20, 25]$ epoch. Weight decay is $5e-4$ and is only applied on $\boldsymbol{\mu}_1$ of the Bayesian layer. The variational parameters $(\boldsymbol{\mu}_1, \boldsymbol{\rho}_1)$ in the Bayesian layer are initialized by $\boldsymbol{\mu}_1\sim \mathcal{N}(0, 0.1)$ and $\mathcal{\boldsymbol{\rho}}\sim \mathcal{N}(-2.25, 0.1)$, and are updated every one sample for faster convergence. Batch size is set to $128$, and SGD with $\text{momentum}=0.9$ is utilized for both training phases. 

\subsection{Evaluation Metrics}
\textbf{Expected calibration error} (ECE) \citep{guo2017calibration} is a widely-used metric for measuring the calibration of networks: predictions are divided into $M$ intervals bins of equal size according to their probability confidence $\hat{p}$. $B_m$ denotes the set of samples whose probability confidence falls into the interval $I_m=(\frac{m-1}{M}, \frac{m}{M}]$ for $m\in\{1,\ldots,M\}$. The average accuracy and confidence of $B_m$ are defined as
\begin{equation}
\operatorname{acc}\left(B_{m}\right) =\frac{1}{\left|B_{m}\right|} \sum_{i \in B_{m}} \mathbf{1}\left[\hat{y}_{i}=y_{i}\right]
\end{equation}
\begin{equation}
    \operatorname{conf}\left(B_{m}\right) =\frac{1}{\left|B_{m}\right|} \sum_{i \in B_{m}} \hat{p}_{i},
\end{equation}
respectively. 
Then ECE is defined as:
\begin{equation}
\mathrm{ECE}=\sum_{m=1}^{M} \frac{\left|B_{m}\right|}{M} \mid \operatorname{acc}\left(B_{m}\right)-\operatorname{conf}\left(B_{m}\right)|.
\end{equation}
Lower ECE corresponds to better calibration performance: the predictive confidence score reflects the prediction accuracy more faithfully, {\it i.e.}, possesses lower predictive uncertainty.

\textbf{Membership inference attack} \citep{shokri2017membership} is often employed for evaluating the capability of privacy preservation in machine learning models, and we utilize a threshold-based version of it \citep{yeom2018privacy} to evaluate our methods in protecting privacy. Given a dataset $\mathcal{S}=(\mathcal{S}_{\text{train}}, \mathcal{S}_{\text{test}})$, where $\mathcal{S}_{\text{train}}$ and $\mathcal{S}_{\text{test}}$ are training set and test set, respectively.

Suppose the ``to-be-inferenced'' example $(\mathbf{x},y)$ comes from $\mathcal{S}$, then the accuracy of membership inference attack with a threshold $\zeta$ is calculated as follow,
\begin{equation}
\begin{aligned}
\operatorname{Acc}(\zeta) &= \frac{1}{2} \times \left( \frac{\sum_{(\mathbf{x},y) \in \mathcal{S}_{\text{train}}} \bm{1}[\hat{p}\left(\mathbf{x}\right) \geq \zeta]}{\vert \mathcal{S}_{\text{train}} \vert}  + \frac{\sum_{(x,y) \in \mathcal{S}_{\text{test}}} \bm{1}[\hat{p}\left(\mathbf{x}\right) < \zeta]}{\vert \mathcal{S}_{\text{test}} \vert} \right).
\end{aligned}
\end{equation}
Then, we can find the optimal threshold $\zeta_{\text{optim}}$ with maximizing the attack accuracy, {\it i.e.},
\begin{equation}
\zeta_{\text{optim}} = \arg \max_{\zeta} \operatorname{Acc}(\zeta),
\end{equation}
and $\operatorname{Acc}(\zeta_{\text{optim}})$ is the final attack accuracy for $f_{\boldsymbol{\theta}}$. Lower $\operatorname{Acc}(\zeta_{\text{optim}})$ means the attacker can hardly judge whether an example $\mathbf{x}\in \mathcal{S}$ belongs to $S_{\text{train}}$ or $S_{\text{test}}$, {\it i.e.}, the attacker infers less information from the model. 

\textbf{Adversarial training} is a popular strategy to enhance the adversarial robustness of neural networks against adversarial examples, which is generated through projected gradient descent (PGD) \citep{madry2018towards} in our empirical experiments. More specifically, adversarial training can be formulated as solving the minimax-loss problem as follow:
\begin{equation}
\min _{\boldsymbol{\theta}} \frac{1}{m} \sum_{i=1}^{m} \max _{\left\|\mathbf{x}_{i}^{\prime}-\mathbf{x}_{i}\right\| \leq \xi} \ell\left(f_{\boldsymbol{\theta}}\left(\mathbf{x}_{i}^{\prime}\right), y_{i}\right),
\end{equation}

where 
$\xi$ is the {\it radius} to limit the distance between adversarial examples and original examples. Adversarial examples are also fed to well-trained models for adversarial attack, and a model is adversarial robust if it has a high accuracy under adversarial attack.

\subsection{Ablation Studies}
\label{sec: layer-wise ablation}

\begin{wrapfigure}{r}{0.46\textwidth}
\vspace{-4mm}
\subfigure[ACC for different layers]{
\begin{minipage}[b]{0.21\textwidth}
            \centering
    		\includegraphics[width=\columnwidth]{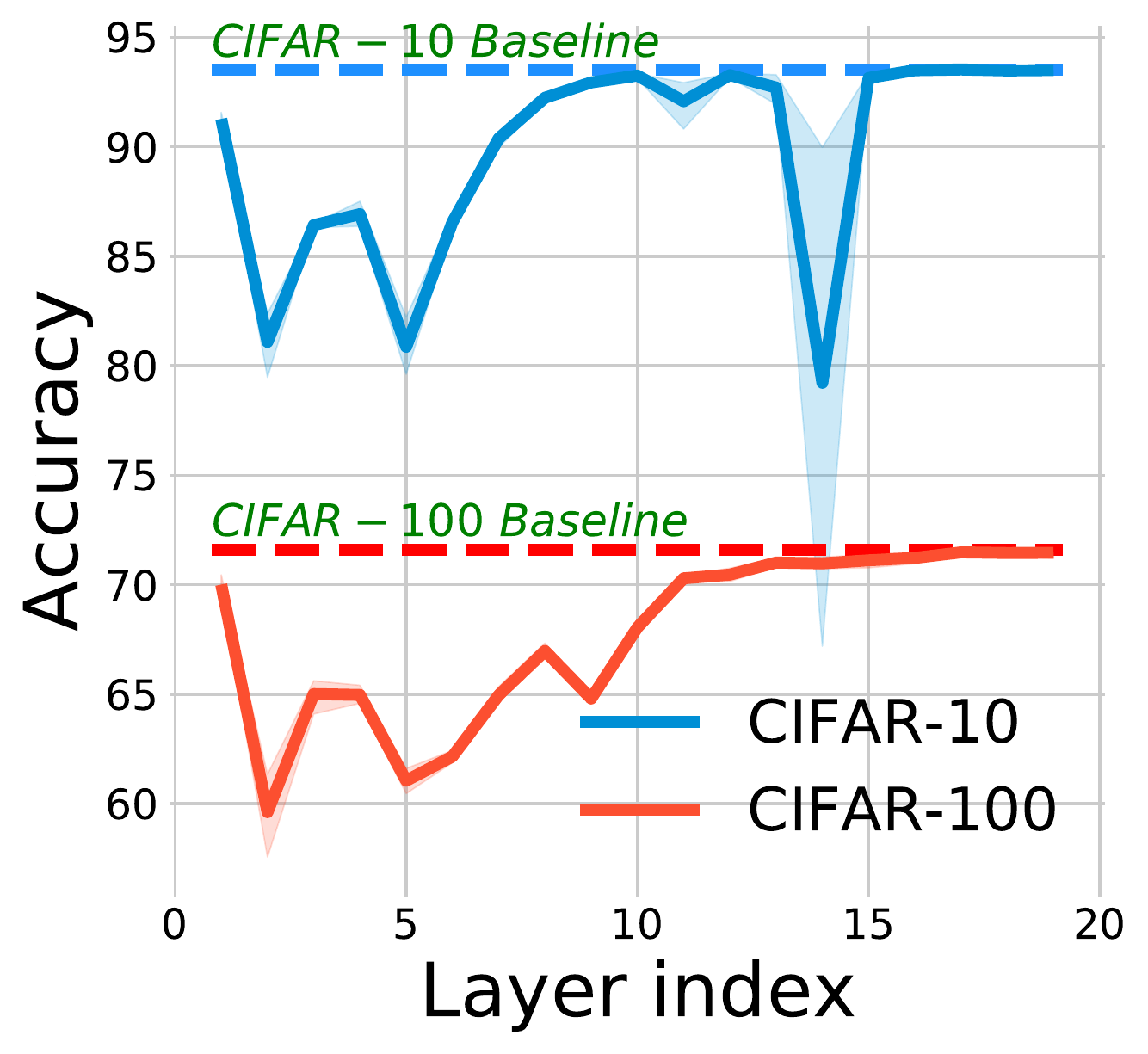}
    		\end{minipage}
		\label{figure:layer wise ablation acc}   
    	}
\subfigure[ECE for different layers]{
\begin{minipage}[b]{0.21\textwidth}
            \centering
   		 	\includegraphics[width=\columnwidth]{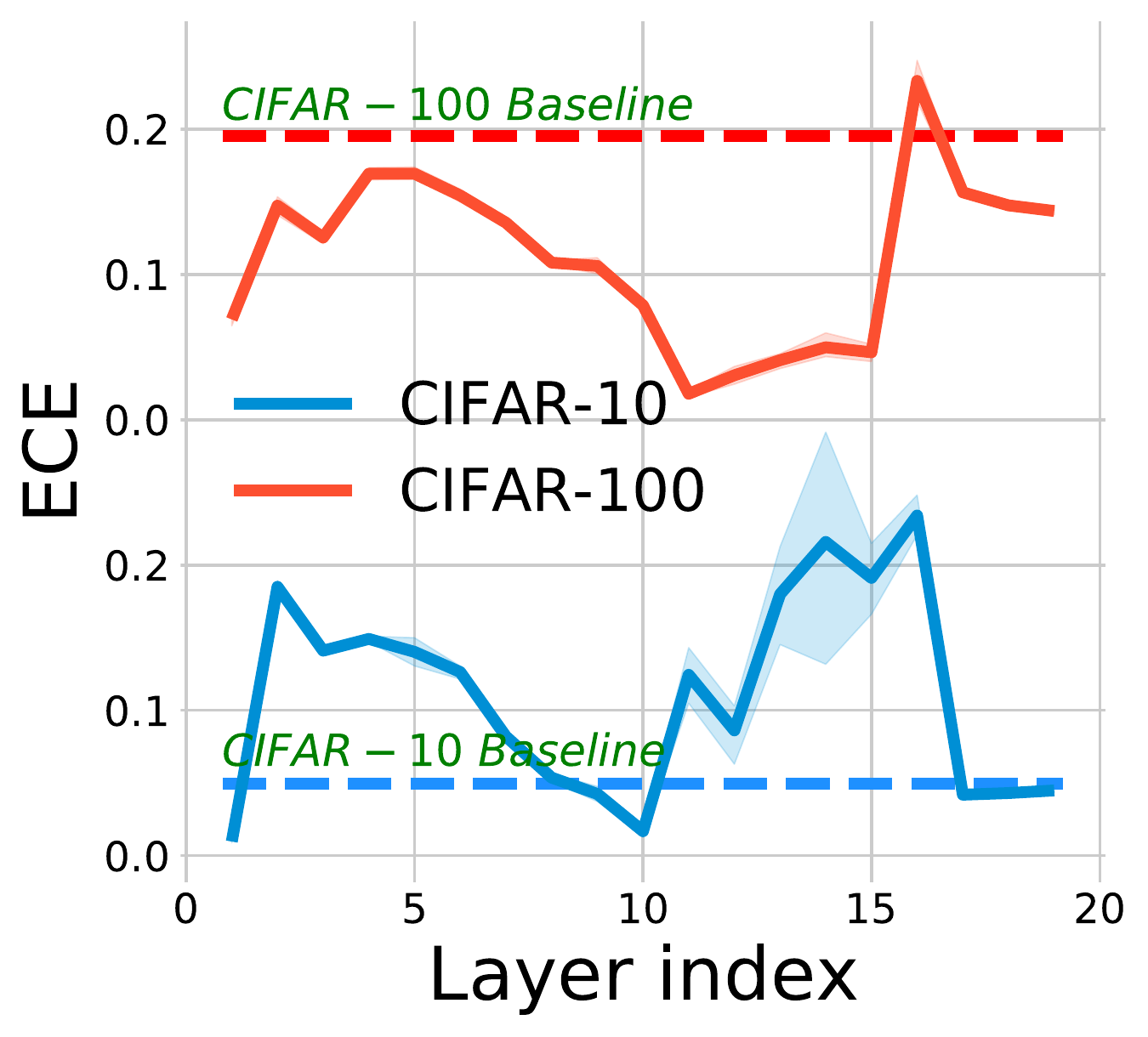}
    		\end{minipage}
		\label{figure:layer wise ablation ece}   
    	}
\vspace{-0.2cm}
\caption{(a) Plot of prediction accuracy formed by varying the layer index of Bayesian layers. (b) Plot of ECE formed by varying the layer index of Bayesian layers. The network architecture is VGG-19, and the models are trained $5$ times with different random seeds.}
\end{wrapfigure}
To explore how the location of Bayesian layers affects the performance of STF-BNNs, we conduct a layer-wise ablation study by specifying different layers as the Bayesian layer retrained in the second training phase. Then the prediction accuracy and ECE performance are measured on these STF-BNNs with Bayesian layers in different layer indexes, as shown in Figure \ref{figure:layer wise ablation acc} and Figure \ref{figure:layer wise ablation ece}. From the figures, we have two observations: (1) there is a good prediction accuracy and a low ECE when the first convolutional layer is converted to the Bayesain layer compared to other layers; (2) according to Figure \ref{figure:layer wise ablation acc}, STF-BNNs can restore predictive ability, which was lost in the initialization of the Bayesian layer, by Bayesian inference in the second training phase; and (3) according to Figure \ref{figure:layer wise ablation ece}, there is a stable drop in ECE in STF-BNNs compared to the baseline, which no longer undergoes the second training phase. 

The layer-wise ablation study verifies the superiority of specifying the first layer as the Bayesian layer, and also shows the potential of STF-BNNs in alleviating overconfidence. Therefore, we adopt the STF-BNN whose first layer is Bayesian, in following experiments.

\subsection{Evaluation on Uncertainty Quantification}

\begin{figure}[t]
\centering  
\subfigure[Confidence threshold vs. Accuracy]{
\begin{minipage}[b]{0.48\textwidth}
            \centering
    		\includegraphics[width=0.48\columnwidth]{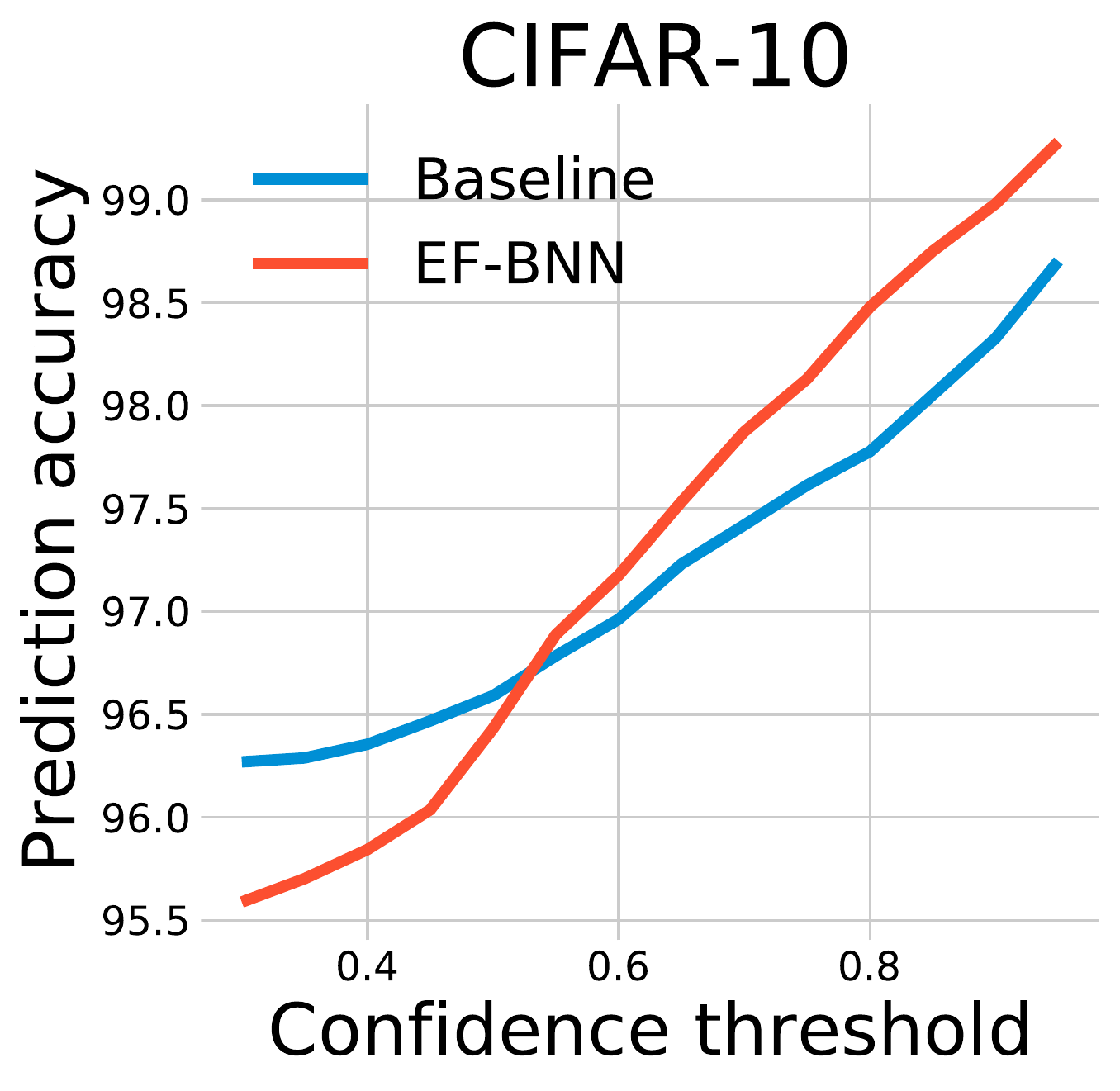}
   		 	\includegraphics[width=0.48\columnwidth]{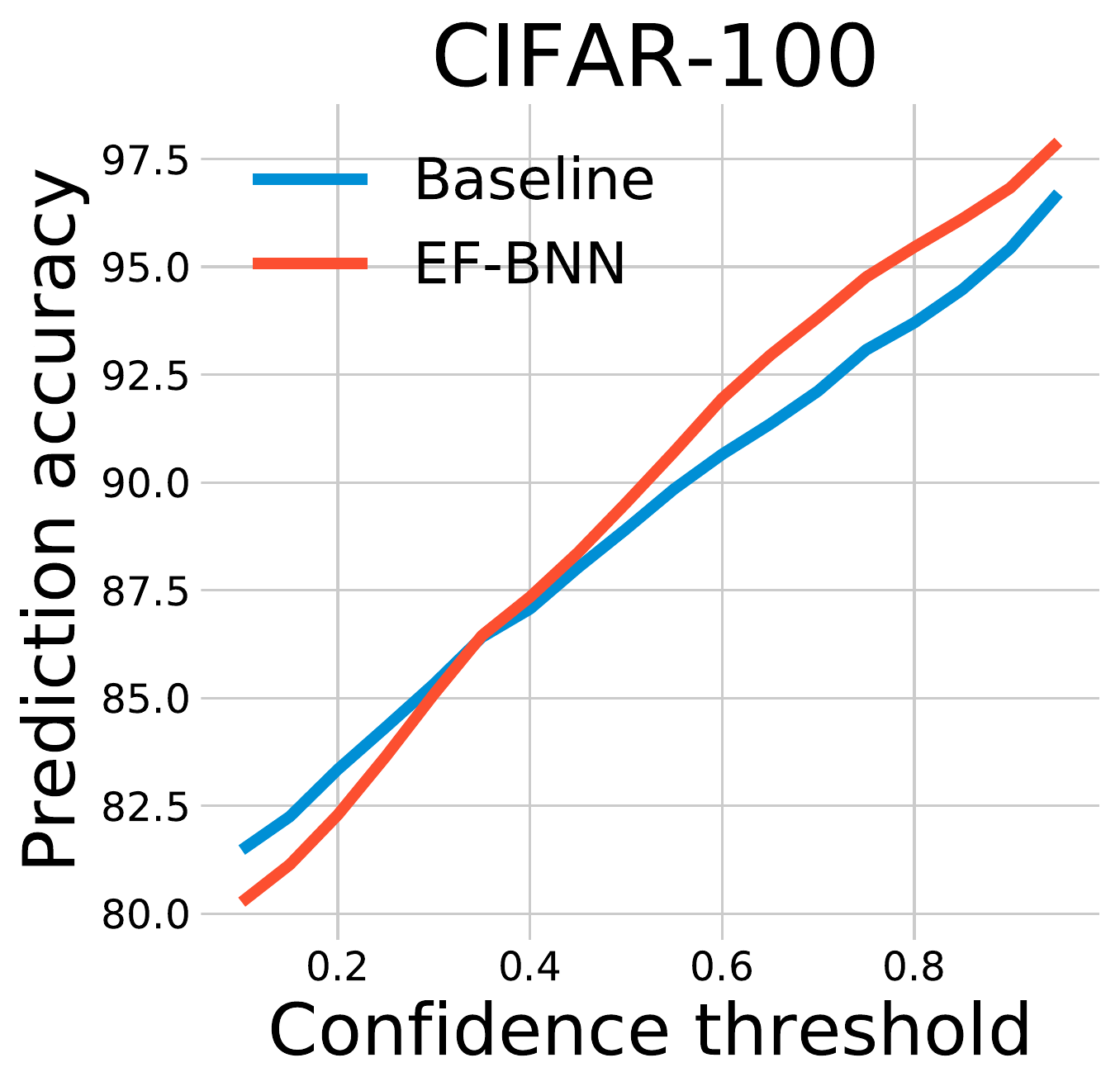}
   		 	\label{figure:acc conf}
    		\end{minipage}
    	}
\subfigure[Accuracy vs. ECE]{
\begin{minipage}[b]{0.48\textwidth}
            \centering
    		\includegraphics[width=0.48\columnwidth]{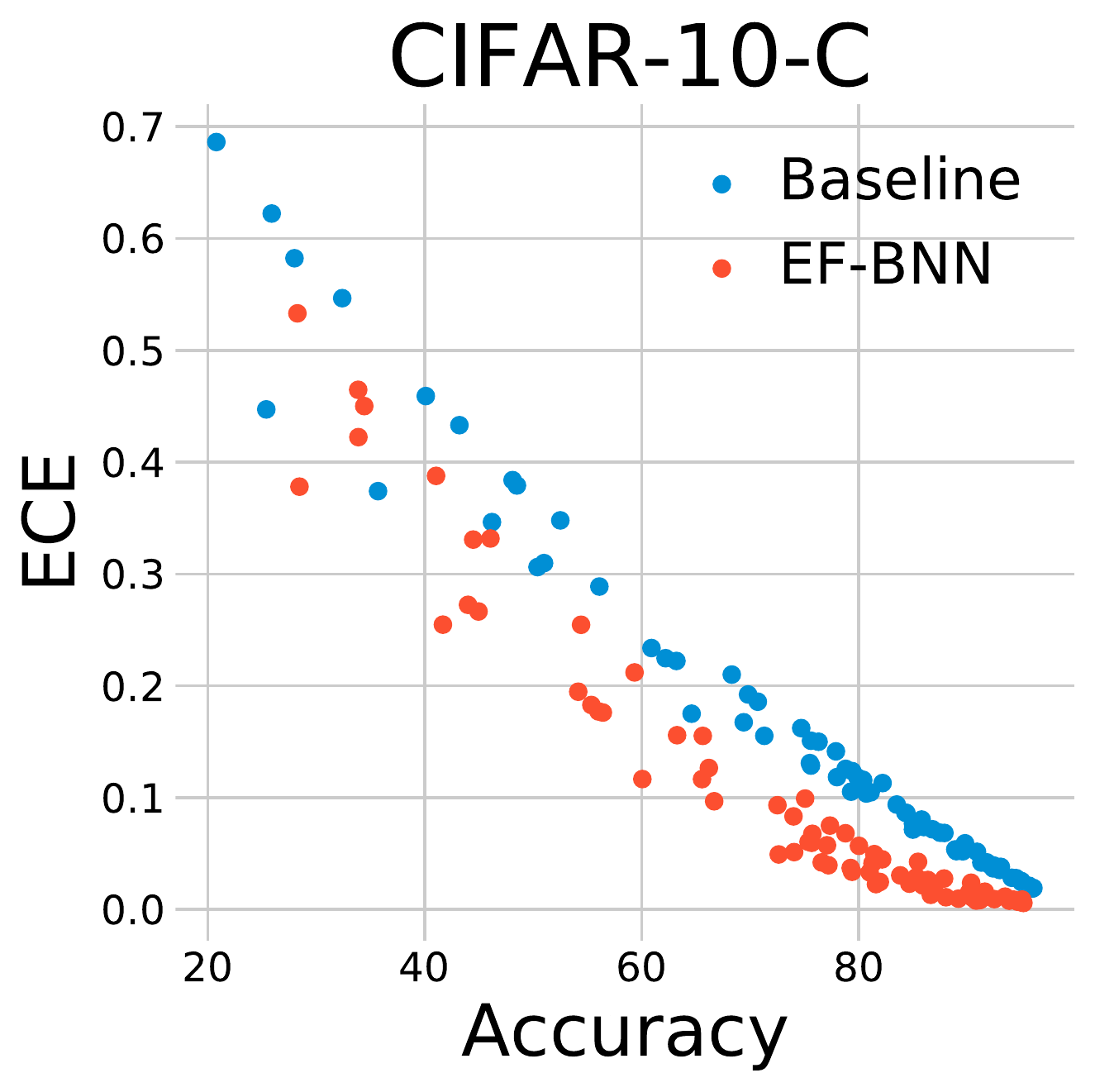}
   		 	\includegraphics[width=0.48\columnwidth]{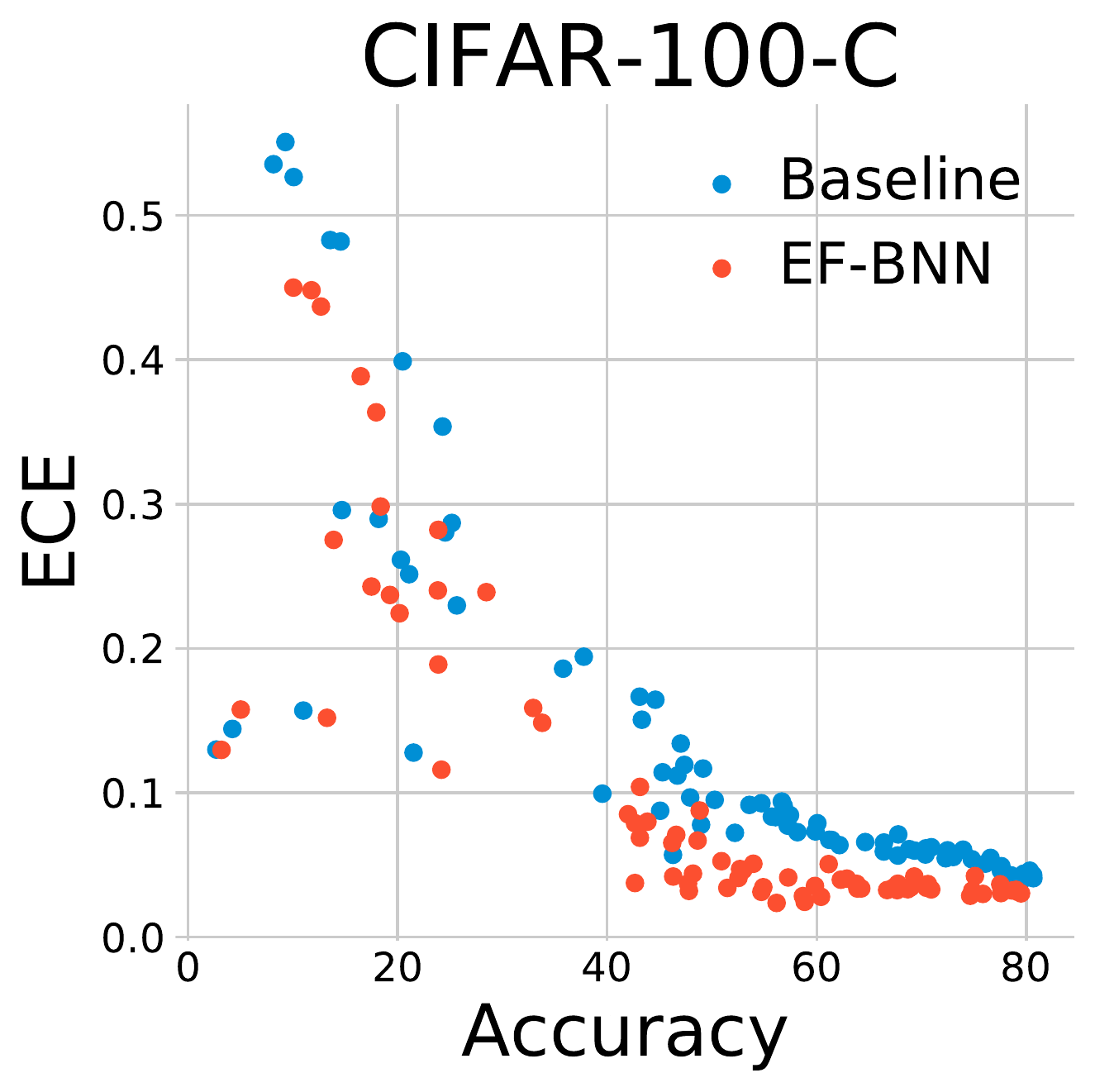}
    		\end{minipage}
		\label{figure:acc_cifar10}   
    	}
\vspace{-0.2cm}
\caption{(a) Test accuracy as a function of confidence threshold on CIFAR-10 (left panel) and CIFAR-100 (right panel). (b) Scatter plots between accuracy and ECE formed by employing STF-BNNs and baseline models on different sub data sets in the corruption data set CIFAR-10 (left panel) and CIFAR-100 (right panel). The network architecture is Wide ResNet-28-10.}
\end{figure}

\textbf{Compared with baseline.} 
We first evaluate the performance under high condidence score with STF-BNNs and baselines on CIFAR-10 and CIFAR-100. To achieve the comparison, we allow models to refuse to give a prediction if the maximum confidence score $\max_i o^{(i)}(f_{\boldsymbol{\theta}}(\mathbf{x}))$ is below a certain confidence threshold, and pictures in Figure \ref{figure:acc conf} are plotted by varying the confidence threshold. From Figure \ref{figure:acc conf}, we observe that when the confidence threshold is large, STF-BNNs attain higher prediction accuracy. In other words, our method is more reliable in the case of a high predictive confidence score compared with the baseline.

We also compare STF-BNNs and the baselines in prediction accuracy and ECE on the corrupted data sets CIFAR-10-C and CIFAR-100-C, each of which contains $75$ sub-datasets with different corruption types and intensities. The comparison result is shown in Figure \ref{figure:acc_cifar10}, and each point in the pictures denotes a sub-dataset of the corruption set. From pictures in Figure \ref{figure:acc_cifar10}, we have an observation that STF-BNN has lower ECE on the corruption sets than the baseline.
\begin{figure}[ht]
\centering  
\subfigure{
\begin{minipage}[b]{0.95\textwidth}
            \centering
   		 	\includegraphics[width=\columnwidth]{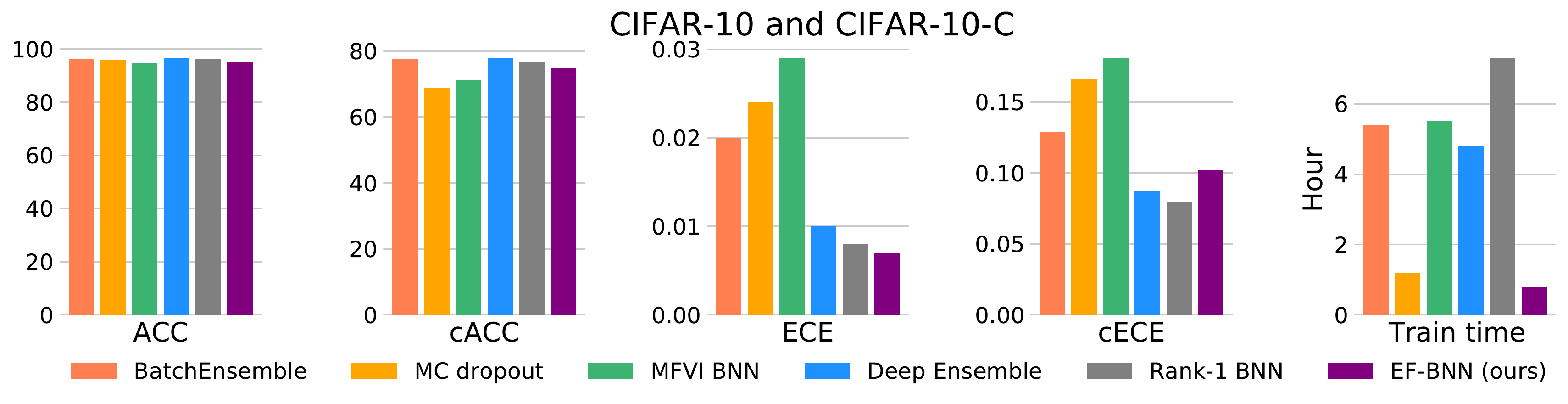}
    		\end{minipage}
    		}
\subfigure{
\begin{minipage}[b]{0.95\textwidth}
            \centering
   		 	\includegraphics[width=\columnwidth]{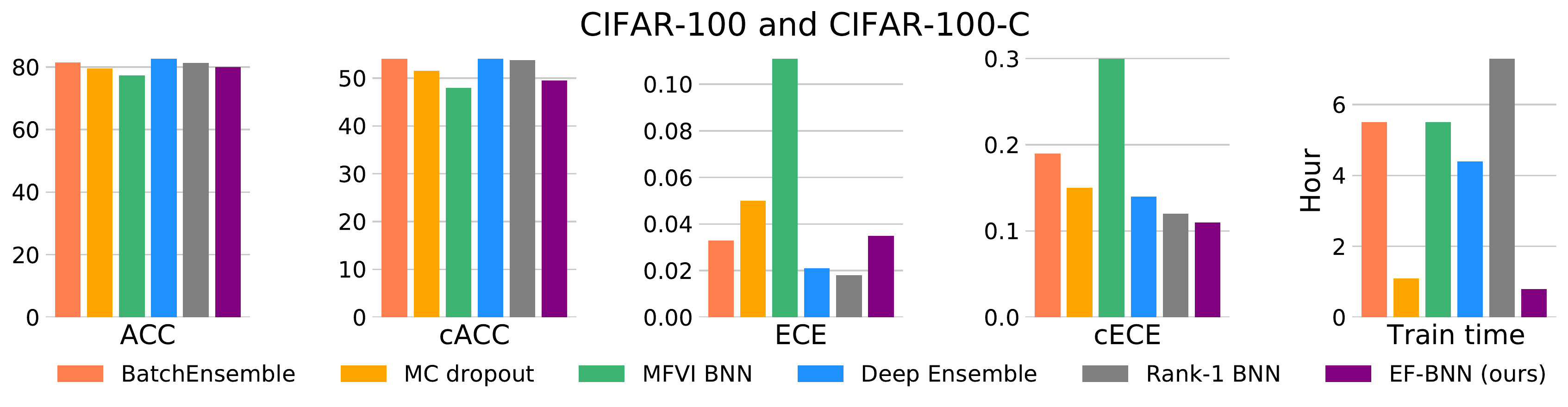}
    		\end{minipage}
    	}
\caption{Plots of comparison between different methods on CIFAR-10 and CIFAR-10-C (top panel), CIFAR-100 and CIFAR-100-C (bottom panel), respectively. cACC and cECE denote accuracy and ECE on CIFAR-10/100-C, respectively. The purple bar represents STF-BNN, which has a decent performance on ECE and cECE and are advantageous on training time. All the methods are based on Wide ResNet-28-10 and averaged over $10$ seeds.}
\label{figure: wrn comparision}
\end{figure}

\textbf{Compared with other methods.} We compared STF-BNNs to other common BNNs or ensembled methods on uncertainty quantification. The result has been shown in Figure \ref{figure: wrn comparision}, and cACC and cECE in the figure denote test accuracy and ECE on CIFAR-10-C and CIFAR-100-C, respectively. From the pictures we obtain the following observations: (1) STF-BNNs (purple bar) have on-par performance of ECE and cECE with the state-of-the-art approaches; (2) training an STF-BNN is time-efficient compared with other methods\footnote{The STF-BNN is trained on $1$ Tesla V100 GPU, and others are trained on $8$ TPUv2 cores according to \href{https://github.com/google/uncertainty-baselines/tree/7af704233ca3905c40171a93e4eeb05c9d8ed608/baselines/cifar}{uncertainty-baselines}. $8$ TPUv2 cores have stronger performance on training neural networks compare with $1$ Tesla V100 GPU.}; (3) MC dropout (yellow bar) also has an short training time, but its performance is significant lower than our method in most cases; and (4) MFVI BNN (green bar), which is the vanilla BNN that only contains random weights and is trained from scratch, has relatively poor performance on both prediction accuracy and ECE, while it needs much longer training time compared to our approach. Therefore, the failure of MFVI BNN 
confirms the effectiveness of our spacial-temporal fusion from the opposite side.

\begin{wrapfigure}{r}{0.45\textwidth}
\vspace{-0.9cm}
\subfigure[Adversarial attack]{
\begin{minipage}[b]{0.21\textwidth}
            \centering
    		\includegraphics[width=1\columnwidth]{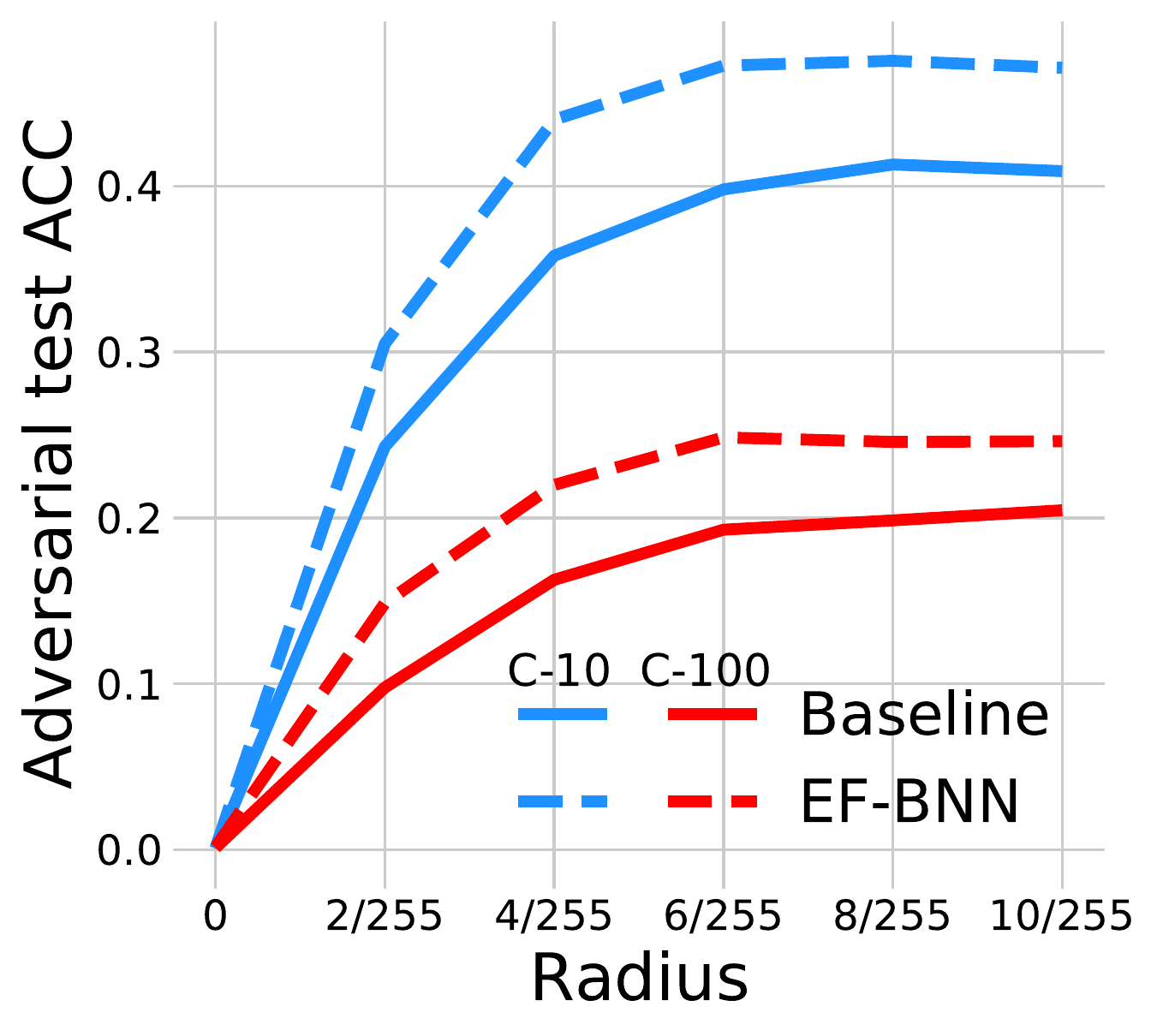}
    		\end{minipage}
    		\label{fig:robustness}
    	}
\subfigure[Member inference attack]{
\begin{minipage}[b]{0.21\textwidth}
            \centering
    		\includegraphics[width=1\columnwidth]{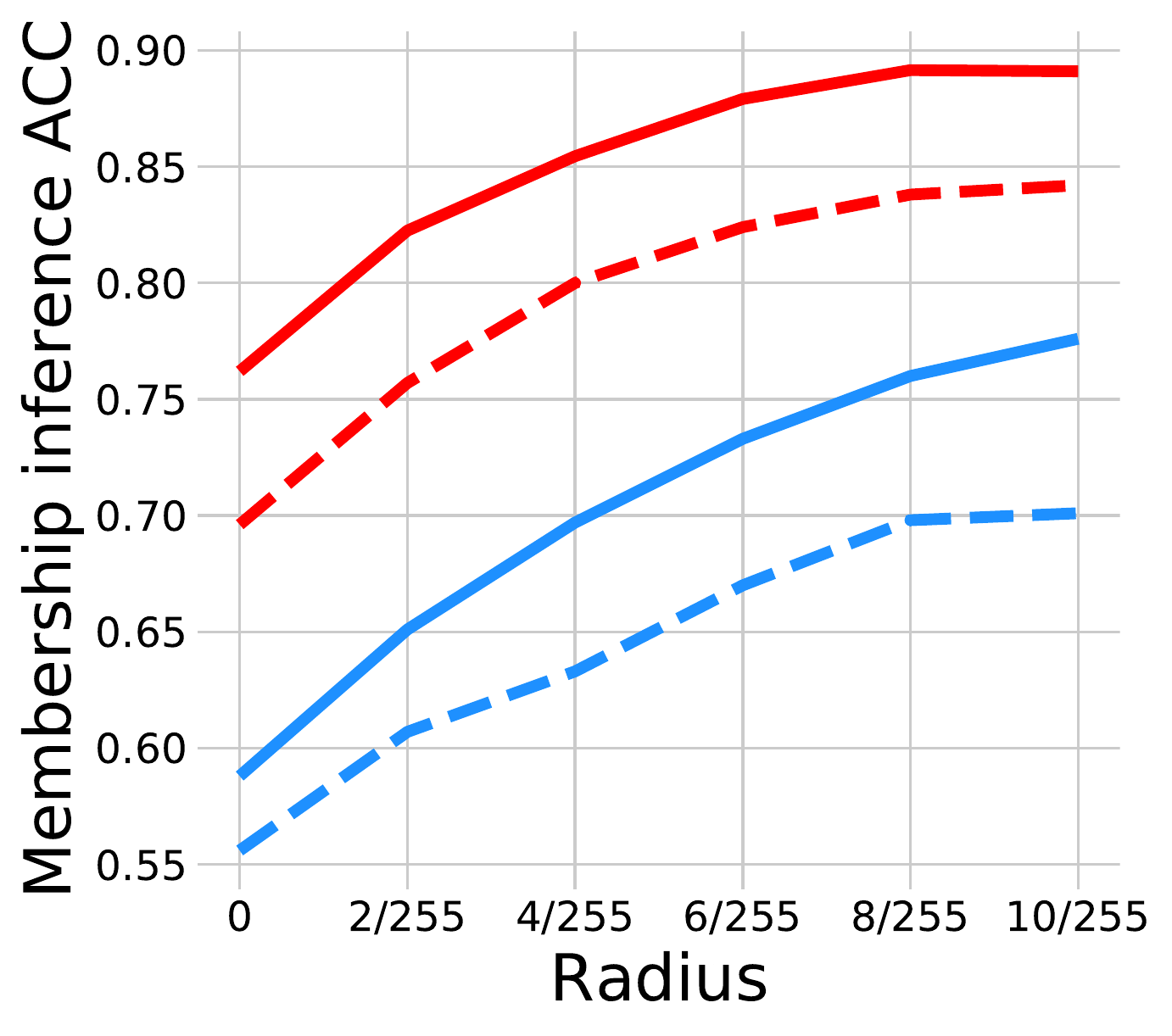}
    		\end{minipage}
    		\label{fig:privacy}
    	}
\vspace{-0.5cm}
\caption{(a) Test accuracy under adversarial attack are function of adversarial radius on CIFAR-10 (C-10) and CIFAR-100 (C-100). (b) Membership inference attack accuracy as a function of adversarial radius. (a) and (b) share the same legend. The radius of test adversarial examples in (a) is set to $10/255$. The network architecture is Wide ResNet-28-10.}
\vspace{-0.2cm}
\end{wrapfigure}
\subsection{Evaluation on Robustness and Privacy}
\label{sec:privacy}
In this section, we evaluate the performance of STF-BNNs on adversarial robustness and privacy preservation through measuring the adversarial attack accuracy and membership inference attack accuracy.

To explore the impact of STF-BNNs on robustness, we deploy the STF-BNN to six well-trained models, which are adversarially trained with six different adversarial radius $\xi$ of $[0, 2/255, 4/255, 6/255, 8/2555, 10/255]$, respectively. Then, we measure the adversarial attack accuracy by employing the adversarial examples with the radius of $10/255$ to attack the adversarial-trained STF-BNNs and corresponding baselines, as shown in Figure \ref{fig:robustness}. From the figure, we observe that our method achieves higher accuracy, {\it i.e.}, preserve better robustness, under adversarial attacks compared to the baselines.

As for the aspect of privacy protection, it is critical for widespread use of machine learning models, but usually exhibits a trade-off with adversarial robustness in adversarial training, see \citet{song2019privacy}. To determine the capability of privacy preservation of our method, 
we employ a membership inference attack on the above adversarial-trained STF-BNNs and corresponding baselines. From the result shown in Figure \ref{fig:privacy}, we obtain an observation that STF-BNNs decrease the membership inference attack accuracy, {\it i.e.}, improve the privacy preserving ability on networks with diverse degrees of robustness. The results suggest that our method helps to eliminate the privacy-preserving compromise due to adversarial training.

\section{Proofs}
The section collects detailed proofs of the results that are omitted in Section \ref{sec:methods}. To avoid technicalities, the measurability/integrability issues are ignored throughout this paper. Moreover, Fubini's theorem is assumed to be applicable for any integration {\it w.r.t.} multiple variables. In other words, the order of integrations is exchangeable. Besides, the norm $\|\cdot\|$ denotes the $l_2$ norm in the following.

\subsection{Proof of generalization bound on STF-BNN}
\label{sec:proof1}
In this section, we present the detailed proof of Theorem 1 based on the PAC-Bayesian framework \citep{mcallester1999pac,mcallester1999some,he2019control}. From the PAC-Bayesian view, a posterior distribution $\mathbb{Q}(\boldsymbol{\theta})$, other than the parameters $\boldsymbol{\theta}_{\operatorname{MLE}}$, is returned by a stochastic learning algorithm. Then, a classic result uniformly bounding the expected risk $\mathcal{R}(\mathbb{Q})$ in terms of the empirical risk $\hat{\mathcal{R}}(\mathbb{Q})$ and the KL divergence $\operatorname{KL}(\mathbb{Q}\|\mathbb{P})$ is as follows.
\begin{lemma}[\citep{mcallester1999pac}, Theorem 2]
\label{lemma:pac}
For any positive real $\delta \in (0,1)$, with probability at least $1-\delta$ over the sample of size $m$, we have the following inequality for all distribution $\mathbb{Q}$:
\begin{equation}
\mathcal{R}(\mathbb{Q}) \leq \hat{\mathcal{R}}(\mathbb{Q})+\sqrt{\frac{\operatorname{KL}(\mathbb{Q} \| \mathbb{P})+\log \frac{1}{\delta}+\log m+2}{2 m-1}},
\end{equation}
where $\operatorname{KL}(\mathbb{Q}\|\mathbb{P})$ is the KL divergence between the distributions $\mathbb{Q}$ and $\mathbb{P}$ and is defined as,
\begin{equation}
\operatorname{KL}(\mathbb{Q}\|\mathbb{P})=\mathbb{E}_{\boldsymbol{\theta} \sim \mathbb{Q}}\left(\log \frac{\mathbb{Q}(\boldsymbol{\theta})}{\mathbb{P}(\boldsymbol{\theta})}\right).
\label{eq:kl}
\end{equation}
\end{lemma}

With the above lemma, we can derive the generalization bound of our STF-BNN.
\begin{proof}[Proof of Theorem 1]
For the STF-BNN, the parameters $\boldsymbol{\theta}$ are partitioned into ${\boldsymbol{\theta}}_1$ and ${\boldsymbol{\theta}}_2$ according to whether they are updated in Bayesian inference or optimization, respectively. We use the standard Gaussian distribution $\mathcal{N}(\boldsymbol{0},I)$ as the priors of ${\boldsymbol{\theta}}_1$ and ${\boldsymbol{\theta}}_2$, which are denoted by $\mathbb{P}(\boldsymbol{\theta}_1)$ and $\mathbb{P}(\boldsymbol{\theta}_2)$.  For ${\boldsymbol{\theta}}_1\sim\mathcal{N}(\boldsymbol{\mu}_1,\boldsymbol{\Sigma}_1)$, we assume the approximate posterior $\mathbb{Q}(\boldsymbol{\theta}_1)= \mathcal{N}(\boldsymbol{\mu}_1,\boldsymbol{\Sigma}_1)$ obeys mean-field assumption, which implies ${\boldsymbol{\Sigma}}_{1}$ is a diagonal matrix. On the other hand, $\boldsymbol{\theta}_2$ is optimized by SGD, which is well-known as the Ornstein-Uhlenbeck process \citep{uhlenbeck1930theory}, and thus $\boldsymbol{\theta}_2$ has an analytic stationary distribution according to \citep{mandt2017stochastic}:
\begin{equation}
\mathbb{Q}(\boldsymbol{\theta}_2)=\frac{1}{\sqrt{2 \pi \operatorname{det}(\boldsymbol{\Sigma}_{2})}} \exp \left\{-\frac{1}{2} \boldsymbol{\theta}_2^{\top} \boldsymbol{\Sigma}_{2}^{-1} \boldsymbol{\theta}_2\right\}.
\end{equation}
For simplify, we denote $\mathbb{Q}(\boldsymbol{\theta}_1)$, $\mathbb{P}(\boldsymbol{\theta}_1)$, $\mathbb{Q}(\boldsymbol{\theta}_2)$, $\mathbb{P}(\boldsymbol{\theta}_2)$ as $\mathbb{Q}_1$, $\mathbb{P}_1$, $\mathbb{Q}_2$, $\mathbb{P}_2$, respectively. Recall $\operatorname{KL}(\mathbb{Q}\|\mathbb{P})$ in Eq. \ref{eq:kl}, we have
\begin{equation}
\begin{aligned}
& \log \frac{\mathbb{Q}(\boldsymbol{\theta})}{\mathbb{P}(\boldsymbol{\theta})} \\
=& \log \frac{\mathbb{Q}(\boldsymbol{\theta}_1, \boldsymbol{\theta}_2)}{\mathbb{P}(\boldsymbol{\theta}_1, \boldsymbol{\theta}_2)} \\
= & \log \frac{\mathbb{Q}(\boldsymbol{\theta}_1)\mathbb{Q}(\boldsymbol{\theta}_2)}{\mathbb{P}(\boldsymbol{\theta}_1)\mathbb{P}(\boldsymbol{\theta}_2)} + \log \frac{\mathbb{Q}(\boldsymbol{\theta}_1, \boldsymbol{\theta}_2)}{\mathbb{Q}(\boldsymbol{\theta}_1)\mathbb{Q}(\boldsymbol{\theta}_2)} \\
= & \log \frac{\mathbb{Q}(\boldsymbol{\theta}_1)}{\mathbb{P}(\boldsymbol{\theta}_1)} + \log \frac{\mathbb{Q}(\boldsymbol{\theta}_2)}{\mathbb{P}(\boldsymbol{\theta}_2)} + \log \frac{\mathbb{Q}(\boldsymbol{\theta}_1, \boldsymbol{\theta}_2)}{\mathbb{Q}(\boldsymbol{\theta}_1)\mathbb{Q}(\boldsymbol{\theta}_2)}.
\end{aligned}
\end{equation}
Hence, $\operatorname{KL}(\mathbb{Q} \| \mathbb{P})$ can be written as:
\begin{equation}
\label{eq:disentangle kl}
    \operatorname{KL}(\mathbb{Q}_1\|\mathbb{P}_1) + \mathbb{E}_{\boldsymbol{\theta} \sim \mathbb{Q}}\left(\log \frac{\mathbb{Q}(\boldsymbol{\theta}_2)}{\mathbb{P}(\boldsymbol{\theta}_2)} + \log \frac{\mathbb{Q}(\boldsymbol{\theta}_1, \boldsymbol{\theta}_2)}{\mathbb{Q}(\boldsymbol{\theta}_1)\mathbb{Q}(\boldsymbol{\theta}_2)}\right).
\end{equation}
By plugging $\mathbb{P}(\boldsymbol{\theta}_2)$ and $\mathbb{Q}(\boldsymbol{\theta}_2)$ into $\log \frac{\mathbb{Q}(\boldsymbol{\theta}_2)}{\mathbb{P}(\boldsymbol{\theta}_2)}$,
\begin{equation}
\label{eq:term 2}
\begin{aligned}
& \log \left(\frac{\mathbb{Q}(\boldsymbol{\theta}_2)}{\mathbb{P}(\boldsymbol{\theta}_2)}\right) \\
=& \log \left(\frac{\sqrt{2 \pi \operatorname{det}(I)}}{\sqrt{2 \pi \operatorname{det}(\boldsymbol{\Sigma}_{2})}} \exp \left\{\frac{1}{2} \boldsymbol{\theta}_2^{\top} I \boldsymbol{\theta}_2-\frac{1}{2} \boldsymbol{\theta}_2^{\top} \boldsymbol{\Sigma}_{2}^{-1} \boldsymbol{\theta}_2\right\}\right) \\
=& \frac{1}{2} \log \left(\frac{1}{\operatorname{det}(\boldsymbol{\Sigma}_{2})}\right)+\frac{1}{2}\left(\boldsymbol{\theta}_2^{\top} I \boldsymbol{\theta}_2-\boldsymbol{\theta}_2^{\top} \boldsymbol{\Sigma}_{2}^{-1} \boldsymbol{\theta}_2\right) .
\end{aligned}
\end{equation}
For the joint posterior distribution $\mathbb{Q}(\boldsymbol{\theta})=\mathbb{Q}(\boldsymbol{\theta}_1,\boldsymbol{\theta}_2)$, its covariance matrix $\boldsymbol{\Sigma}=\left[\begin{array}{ll}
\boldsymbol{\Sigma}_{1} & \boldsymbol{\Sigma}_{12} \\
\boldsymbol{\Sigma}_{12}^{\top} & \boldsymbol{\Sigma}_{2}
\end{array}\right]$. However, for $\mathbb{Q}(\boldsymbol{\theta}_1)\mathbb{Q}(\boldsymbol{\theta}_2)$, because ${\boldsymbol{\theta}}_1$ and ${\boldsymbol{\theta}}_2$ are independent in this distribution, $\boldsymbol{\Sigma}_{12}=\operatorname{cov}(\boldsymbol{\theta}_1,\boldsymbol{\theta}_2)=\boldsymbol{0}$ and its covariance matrix $\boldsymbol{\Sigma}^\prime=\left[\begin{array}{ll}
\boldsymbol{\Sigma}_{1} & \boldsymbol{0} \\
\boldsymbol{0} & \boldsymbol{\Sigma}_{2}
\end{array}\right]$. Therefore, $\log \frac{\mathbb{Q}(\boldsymbol{\theta}_1, \boldsymbol{\theta}_2)}{\mathbb{Q}(\boldsymbol{\theta}_1)\mathbb{Q}(\boldsymbol{\theta}_2)}$ can be written as
\begin{equation}
\label{eq:term 3}
\begin{aligned}
& \log \frac{\mathbb{Q}(\boldsymbol{\theta}_1, \boldsymbol{\theta}_2)}{\mathbb{Q}(\boldsymbol{\theta}_1)\mathbb{Q}(\boldsymbol{\theta}_2)} \\
=& \log \left(\frac{\sqrt{2 \pi \operatorname{det}(\boldsymbol{\Sigma}^\prime)}}{\sqrt{2 \pi \operatorname{det}(\boldsymbol{\Sigma})}} \exp \left\{\frac{1}{2} \boldsymbol{\theta}^{\top} \boldsymbol{\Sigma}^\prime \boldsymbol{\theta}-\frac{1}{2} \boldsymbol{\theta}^{\top} \boldsymbol{\Sigma}^{-1} \boldsymbol{\theta}\right\}\right) \\
=& \frac{1}{2} \log \left(\frac{\operatorname{det}(\boldsymbol{\Sigma}^\prime)}{\operatorname{det}(\boldsymbol{\Sigma})}\right)+\frac{1}{2}\left(\boldsymbol{\theta}^{\top} \boldsymbol{\Sigma}^\prime \boldsymbol{\theta} -\boldsymbol{\theta}^{\top} \boldsymbol{\Sigma}^{-1} \boldsymbol{\theta}\right) .
\end{aligned}
\end{equation}
By plugging Eq. \ref{eq:term 2} and \ref{eq:term 3} into Eq. \ref{eq:disentangle kl}, we have
\begin{equation}
\begin{aligned}
& \mathbb{E}_{\boldsymbol{\theta} \sim \mathbb{Q}}\left(\log \frac{\mathbb{Q}(\boldsymbol{\theta}_2)}{\mathbb{P}(\boldsymbol{\theta}_2)} + \log \frac{\mathbb{Q}(\boldsymbol{\theta}_1, \boldsymbol{\theta}_2)}{\mathbb{Q}(\boldsymbol{\theta}_1)\mathbb{Q}(\boldsymbol{\theta}_2)}\right) \\
=&  \int_{\boldsymbol{\theta} \in \Theta} \left[\frac{1}{2} \log \left(\frac{1}{\operatorname{det}(\boldsymbol{\Sigma}_{2})}\right)+\frac{1}{2}\left(\boldsymbol{\theta}_2^{\top} I \boldsymbol{\theta}_2-\boldsymbol{\theta}_2^{\top} \boldsymbol{\Sigma}_{2}^{-1} \boldsymbol{\theta}_2\right)\right. \\
& + \left.\frac{1}{2} \log \left(\frac{\operatorname{det}(\boldsymbol{\Sigma}^\prime)}{\operatorname{det}(\boldsymbol{\Sigma})}\right)+\frac{1}{2}\left(\boldsymbol{\theta}^{\top} \boldsymbol{\Sigma}^\prime \boldsymbol{\theta} -\boldsymbol{\theta}^{\top} \boldsymbol{\Sigma}^{-1} \boldsymbol{\theta}\right) \right] q(\boldsymbol{\theta}) \mathrm{d} \boldsymbol{\theta} \\
=&   \frac{1}{2}\log \left(\frac{\operatorname{det}(\boldsymbol{\Sigma}_{1})\operatorname{det}(\boldsymbol{\Sigma}_{2})}{\operatorname{det}(\boldsymbol{\Sigma})\operatorname{det}(\boldsymbol{\Sigma}_{2})}\right) + \frac{1}{2} \mathbb{E}_{\boldsymbol{\theta}_2} \left[\boldsymbol{\theta}_2^{\top} I \boldsymbol{\theta}_2\right] - \\
& \frac{1}{2} \mathbb{E}_{\boldsymbol{\theta}} \left[\boldsymbol{\theta}_2^{\top} \boldsymbol{\Sigma}_{2} \boldsymbol{\theta}_2\right]  + \frac{1}{2} \mathbb{E}_{\boldsymbol{\theta}} \left[\boldsymbol{\theta}^{\top} \boldsymbol{\Sigma}^\prime \boldsymbol{\theta}\right] - \frac{1}{2} \mathbb{E}_{\boldsymbol{\theta}} \left[\boldsymbol{\theta}^{\top} I \boldsymbol{\theta}\right] \\
=&   \frac{1}{2}\log \left(\frac{\operatorname{det}(\boldsymbol{\Sigma}_{1})}{\operatorname{det}(\boldsymbol{\Sigma})}\right) + \frac{1}{2}\operatorname{tr}(\boldsymbol{\Sigma}_{2}-I) + \frac{1}{2}\operatorname{tr}(\boldsymbol{\Sigma}^\prime\boldsymbol{\Sigma}-I) \\
\leq&   \frac{1}{2}\operatorname{tr}(\boldsymbol{\Sigma}_{1}-I) - \frac{1}{2}\log \operatorname{det}(\boldsymbol{\Sigma})  + \frac{1}{2}\operatorname{tr}(\boldsymbol{\Sigma}_{2}-I) + \frac{1}{2}\operatorname{tr}(\boldsymbol{\Sigma}^\prime\boldsymbol{\Sigma}-I) \\
=&  - \frac{1}{2}\log \operatorname{det}(\boldsymbol{\Sigma}) + \frac{1}{2}\operatorname{tr}(\boldsymbol{\Sigma}-I) + \frac{1}{2}\operatorname{tr}(\boldsymbol{\Sigma}^\prime\boldsymbol{\Sigma}-I) \\
=&  - \frac{1}{2}\log \operatorname{det}(\boldsymbol{\Sigma}) + \frac{1}{2}\operatorname{tr}(\boldsymbol{\Sigma}-2I) + \frac{1}{2}(\|\boldsymbol{\Sigma}_{1}\|^2+\|\boldsymbol{\Sigma}_{2}\|^2)
\end{aligned}
\end{equation}

Then plugging the above equation and Eq. \ref{eq:disentangle kl} into Lemma \ref{lemma:pac} yields the desired inequality and the proof is finished.
\end{proof}

\subsection{Proof of the capability of Mitigating Overconfidence about STF-BNN}
\label{sec:proof2}
In this section, we present the detailed proof of Theorem 2. For a ReLU network, its input space is partitioned into linear regions by the nonlinearities at the activation \citep{pascanu2013number,montufar2014number}. The output-input mapping induced by a ReLU network is linear with respect to the input data within linear regions and nonlinear and non-smooth in the boundaries between linear regions. Then, with the notion of linear regions, we introduced Lemma \ref{lemma:hein relu} as follows.
\begin{lemma}[\citep{hein2019relu}, Lemma 3.1]
\label{lemma:hein relu}
Let $\{Q_i\}_{l-1}^R$ be the set of linear regions associated to the ReLU network $f: \mathbb{R}^n\rightarrow \mathbb{R}^k$. For any $x\in \mathbb{R}^n$ there exists an $\alpha >0$ and $t\in\{1,...,R\}$ such that $\delta x\in Q_{t}$ for all $\delta \geq \alpha$. Furthermore, the restriction of $f$ to $Q_t$ can be written as an affine function $\mathbf{U}x+\mathbf{c}$ for some suitable $\mathbf{U}\in \mathbb{R}^{k\times n}$ and $\mathbf{c}\in \mathbb{R}^k$.
\label{lemma:linear region}
\end{lemma}
With the above lemma, we can prove the capability of mitigating overconfidence of our STF-BNN.

\begin{proof}[Proof of Theorem 2]
The binary STF-BNN $f_{\boldsymbol{\theta}_1, \boldsymbol{\theta}_2}:\mathbb{R}^n\rightarrow\mathbb{R}$ consists of $\boldsymbol{\theta}_1$ and $\boldsymbol{\theta}_2$. 
According to Lemma \ref{lemma:linear region} there exists an $\alpha>0$ and a linear region $R$, along with $\mathbf{u}^\top\in \mathbb{R}^r$, $\mathbf{w}\in \mathbb{R}^{r\times n}$, $\mathbf{b}\in \mathbb{R}^r$, and $c\in \mathbb{R}$, such that for any $\delta \geq \alpha$, we have that $\delta \mathbf{x} \in R$ and restriction $f_{\boldsymbol{\theta}_1,\boldsymbol{\theta}_2|R}$ can be written as $\mathbf{u}^\top(\mathbf{w}\mathbf{x}+\mathbf{b})+{c}$, where $r$ is the output dimension of the first Bayesian layer in the STF-BNN. Therefore for any such $\delta$, we can write the gradient $\mathbf{d}_1(\delta\boldsymbol{x})=\left.\nabla_{\boldsymbol{\theta}_1} \left(\mathbf{u}^\top(\mathbf{w}\delta\mathbf{x}+\mathbf{b})+{c}\right)\right|_{\boldsymbol{\mu}_1}$ as follows:
\begin{equation}
\begin{aligned}
\mathbf{d}(\delta {\mathbf{x}}) &=\left.\frac{\partial\left(\delta \mathbf{u}^\top \mathbf{w}\mathbf{x}\right)}{\partial \boldsymbol{\theta}_1}\right|_{\boldsymbol{\mu}_1}
+\left.\frac{\partial \mathbf{u}^\top\mathbf{b}}{\partial \boldsymbol{\theta}_1}\right|_{\boldsymbol{\mu}_1} + \left.\frac{\partial {c}}{\partial \boldsymbol{\theta}_1}\right|_{\boldsymbol{\mu}_1} \\
&=\left.\delta \frac{\partial \mathbf{u}^\top\mathbf{wx}}{\partial \boldsymbol{\theta}_1}\right|_{\boldsymbol{\mu}_1}{\mathbf{x}}
+\left.\frac{\partial \mathbf{u}^\top\mathbf{b}}{\partial \boldsymbol{\theta}_1}\right|_{\boldsymbol{\mu}_1} \\
&=\delta\left(\mathbf{u}^\top \mathbf{J} {\mathbf{x}}+\frac{1}{\delta} \nabla_{\boldsymbol{\theta}_1} \mathbf{u}^\top\mathbf{b}|_{\boldsymbol{\mu}_1}\right),
\end{aligned}
\end{equation}
where $\mathbf{J}=\left.\frac{\partial \mathbf{w}}{\partial \boldsymbol{\theta}_1}\right|_{\boldsymbol{\mu}_1}$. Then, by Eq. \ref{eq:ef bnn zx} 
we have
\begin{equation}
\begin{aligned}
|z(\delta {\mathbf{x}})|=&\frac{\left|\delta \mathbf{u}^\top\mathbf{w} {\mathbf{x}}+ \mathbf{u}^\top\mathbf{b} + {c}\right|}{\sqrt{1+\pi / 8 \mathbf{d}_1(\delta {\mathbf{x}})^{\top} \mathbf{\Sigma}_1 \mathbf{d}_1(\delta {\mathbf{x}})}} \\
=& \frac{\left|\delta\left(\mathbf{u}^\top\mathbf{w} {\mathbf{x}}+\frac{1}{\delta} \mathbf{u}^\top\mathbf{b}+\frac{1}{\delta}\mathbf{c}\right)\right|}{\sqrt{1+\pi / 8 \delta^{2}\boldsymbol{\Omega}^{\top} \mathbf{\Sigma}_1\boldsymbol{\Omega}}} \\
=& \frac{\left|\left(\mathbf{u}^\top\mathbf{w} {\mathbf{x}}+\frac{1}{\delta} \mathbf{u}^\top\mathbf{b}+\frac{1}{\delta}\mathbf{c}\right)\right|}{\sqrt{\frac{1}{\delta^2}+\pi / 8 \boldsymbol{\Omega}^{\top} \mathbf{\Sigma}_1\boldsymbol{\Omega}}},
\end{aligned}
\end{equation}
where $\boldsymbol{\Omega}=\mathbf{u}^\top \mathbf{J} {\mathbf{x}}+\frac{1}{\delta} \nabla_{\boldsymbol{\theta}_1} \mathbf{u}^\top\mathbf{b}|_{\boldsymbol{\mu}_1}$.
Notice that as $\delta \rightarrow \infty$, $1/\delta$ and $1/\delta^2$ go to $0$. Hence, in the limit, we have that
\begin{equation}
\lim _{\delta \rightarrow \infty}|z(\delta {\mathbf{x}})|=\frac{\left|\mathbf{u}^\top\mathbf{w} {\mathbf{x}}\right|}{\sqrt{\pi / 8\left(\mathbf{u}^\top\mathbf{J}{\mathbf{x}}\right)^{\top} \mathbf{\Sigma}_1\left(\mathbf{u}^\top\mathbf{J} {\mathbf{x}}\right)}}.
\end{equation}
To get the desired result, the following lemmas in terms of matrix operations are needed.
\begin{lemma}[\citep{kristiadi2020being}, Lemma A.2]
\label{lemma:a3}
Let $\mathbf{x}\in \mathbb{R}^n$ and $\mathbf{A}\in \mathbb{R}^{n\times n}$ be an SPD matrix, then $\mathbf{x}^\top\mathbf{Ax}\geq\lambda_{\min}(\mathbf{A})\|\mathbf{x}\|^2.$
\end{lemma}
\begin{lemma}[\citep{kristiadi2020being}, Lemma A.3]
\label{lemma:a4}
Let $\mathbf{A}\in\mathbb{R}^{r\times n}$ and $\mathbf{z}\in \mathbb{R}^n$ with $r\geq n$, then $\|\mathbf{Az}\|^2\geq s_{\min}^2(\mathbf{A})\|\mathbf{z}\|^2$.
\end{lemma}
Then, by using the Cauchy-Schwarz inequality, Lemma \ref{lemma:a3}, and Lemma \ref{lemma:a4} sequentially, we can upper-bound this limit:
\begin{equation}
\begin{aligned}
\lim _{\delta \rightarrow \infty}|z(\delta {\mathbf{x}})| &=\frac{\left|\mathbf{u}^\top\mathbf{w} {\mathbf{x}}\right|}{\sqrt{\pi / 8\left(\mathbf{u}^\top\mathbf{J}{\mathbf{x}}\right)^{\top} \mathbf{\Sigma}_1\left(\mathbf{u}^\top\mathbf{J} {\mathbf{x}}\right)}}\\
&\leq \frac{\|\mathbf{u}^\top\|\|\mathbf{w}\|\|\mathbf{x}\|}{\sqrt{\pi / 8 \lambda_{\min}(\boldsymbol{\Sigma}_1)\|\mathbf{u}^\top\mathbf{Jx}\|^2}} \\
&\leq \frac{\|\mathbf{u}^\top\|\|\mathbf{w}\|}{s_{\min}(\mathbf{u}^\top\mathbf{J})\sqrt{\pi / 8 \lambda_{\min}(\boldsymbol{\Sigma}_1)}},
\end{aligned}
\end{equation}
With $\|\mathbf{u}^\top\|=\|\mathbf{u}\|$, thus the proof is complete.
\end{proof}

\section{Conclusion}
This paper designs the spatial-temporal-fusion BNN, which fuses the optimization and Bayesian inference from both spatial and temporal aspects to efficiently improve the reliability of neural networks. This fusion is based on an empirical finding that the first layer of the network has lower stability when it is retrained. Theoretical analysis has been provided on the generalization ability and the capability of mitigating overconfidence of our STF-BNN. Sufficient empirical studies present that STF-BNNs (1) achieve on-par performance on prediction and uncertainty quantification with the state-of-the-art methods; (2) have a tremendous decrease in training time and required memory; and (3) significantly improve adversarial robustness and privacy preservation.

\bibliographystyle{plainnat}
\bibliography{bnn}

\end{document}